\documentclass{article}

\PassOptionsToPackage{numbers, compress}{natbib}

 \usepackage[preprint]{neurips_2026}

\usepackage[utf8]{inputenc} 
\usepackage[T1]{fontenc}    
\usepackage{hyperref}       

\newcommand{\appref}[1]{\hyperref[#1]{App.~\ref*{#1}}}

\usepackage{url}            
\usepackage{booktabs}       
\usepackage{amsfonts}       
\usepackage{nicefrac}       
\usepackage{microtype}      
\usepackage{xcolor}         
\usepackage{todonotes}      
\usepackage{caption}
\usepackage{subcaption}
\usepackage{wrapfig}
\usepackage{tikz}
\usetikzlibrary{arrows.meta,calc,positioning,fit,backgrounds,shapes.geometric}

\usepackage[most]{tcolorbox} 
\usepackage{soul} 

\usepackage{tabularx}
\usepackage{colortbl}
\definecolor{klettermixrow}{gray}{0.94}

\definecolor{KMBadBack}{HTML}{FFF5F5}
\definecolor{KMBadFrame}{HTML}{B42318}
\definecolor{KMWarnBack}{HTML}{FFF8E1}
\definecolor{KMWarnFrame}{HTML}{B7791F}
\definecolor{KMGoodBack}{HTML}{F2FBF5}
\definecolor{KMGoodFrame}{HTML}{1A7F37}
\definecolor{KMPanelFrame}{HTML}{D0D7DE}
\definecolor{KMPanelTitle}{HTML}{F6F8FA}
\definecolor{KMBadHighlight}{HTML}{F8D7DA}
\definecolor{KMWarnHighlight}{HTML}{FFF3CD}
\definecolor{KMGoodHighlight}{HTML}{D1E7DD}

\newcommand{\badspan}[1]{{\sethlcolor{KMBadHighlight}\hl{#1}}}

\newcommand{\goodspan}[1]{{\sethlcolor{KMGoodHighlight}\hl{#1}}}
\newcommand{\klettermixtokens}{725B} 

\tcbset{
  kmexamplecard/.style={
    enhanced,
    boxrule=0.45pt,
    arc=1.2mm,
    left=6pt,
    right=6pt,
    top=6pt,
    bottom=6pt,
    before skip=0.65em,
    after skip=0.9em
  },
  kmexamplepanel/.style={
    enhanced,
    colback=white,
    colframe=KMPanelFrame,
    colbacktitle=KMPanelTitle,
    coltitle=black,
    boxrule=0.35pt,
    arc=0.9mm,
    left=5pt,
    right=5pt,
    top=4pt,
    bottom=4pt,
    fonttitle=\footnotesize\bfseries,
    before skip=0pt,
    after skip=0pt
  }
}

\newcommand{\translationexample}[6]{%
  \begin{tcolorbox}[kmexamplecard,colback=#1,colframe=#2]
    {\bfseries #3}\par
    {\footnotesize\emph{#4}}\par\vspace{0.45em}
    \begin{tcbraster}[raster columns=2,raster equal height=rows,raster column skip=6pt]
      \begin{tcolorbox}[kmexamplepanel,title={English source}]
        \footnotesize #5
      \end{tcolorbox}
      \begin{tcolorbox}[kmexamplepanel,title={Target output}]
        \footnotesize #6
      \end{tcolorbox}
    \end{tcbraster}
  \end{tcolorbox}
}
\newcommand{\badexample}[4]{\translationexample{KMBadBack}{KMBadFrame}{#1}{#2}{#3}{#4}}
\newcommand{\goodexample}[4]{\translationexample{KMGoodBack}{KMGoodFrame}{#1}{#2}{#3}{#4}}

\title{\datasetname{}: Climbing Toward High-Quality German Pretraining Data -- The Full Report}

\author{%
\begin{minipage}{\dimexpr\textwidth-2\tabcolsep\relax}
\centering
\normalfont
{\bfseries\small
Maurice Kraus\textsuperscript{\normalfont 1,2}\thanks{Equal contribution.}\quad
Ruben H{\"a}rle\textsuperscript{\normalfont 1,2}\footnotemark[1]\quad
Sebastian Sztwiertnia\textsuperscript{\normalfont 1,2}\quad
Abbas Goher Khan\textsuperscript{\normalfont 4}\\[-0.1ex]
Mehdi Ali\textsuperscript{\normalfont 3,4}\quad
Michael Fromm\textsuperscript{\normalfont 3,4}\quad
Nicolas Flores-Herr\textsuperscript{\normalfont 4}\quad
Kristian Kersting\textsuperscript{\normalfont 1,5,6,7}
}\\[0.45em]
{\footnotesize
\textsuperscript{1}AI \& ML Group, TU Darmstadt\quad
\textsuperscript{2}Lab1141\quad
\textsuperscript{3}Lamarr Institute\quad
\textsuperscript{4}Fraunhofer IAIS\\[-0.1ex]
\textsuperscript{5}hessian.AI\quad
\textsuperscript{6}German Research Center for AI (DFKI)\quad
\textsuperscript{7}Centre for Cognitive Science, TU Darmstadt
}\\[0.45em]
{\scriptsize\ttfamily
\{maurice.kraus,ruben.haerle\}@tu-darmstadt.de\\[-0.1ex]
}
\end{minipage}%
}

\newcommand{\datasetname}[0]{KletterMix}

\begin{document}

\maketitle
\setcounter{footnote}{0}

\begin{abstract}
High-quality pretraining data is a central ingredient in modern language models, but German-language resources remain far less developed than their English counterparts: they are often smaller, less carefully curated, weakly documented, and rarely validated through controlled training experiments. We introduce \datasetname{}, a high-quality German corpus for language model pretraining and annealing, designed as a reusable dataset artifact for the natural language processing and modeling community.
\datasetname{} is built by translating a state-of-the-art English pretraining corpus into German while preserving document boundaries, metadata, source structure, and topical diversity. This construction yields a German corpus with the scale and diversity of a modern pretraining dataset, while enabling direct comparison to its English source. We document the dataset through a broad set of corpus-level analyses, including translation quality, document length distributions, topic coverage, source composition, and geographic metadata. Using COMETKiwi, we show that the translated documents achieve strong quality across diverse domains, suggesting that careful translation can preserve much of the semantic and stylistic richness of the original corpus.
Beyond dataset construction, we evaluate \datasetname{} as training data. Through controlled pretraining and annealing ablations against established German corpora, we show that models trained on \datasetname{} achieve measurable improvements on German-language downstream evaluations. 
These results demonstrate that carefully curated translated data can substantially strengthen the German pretraining data ecosystem.\footnotemark
\end{abstract}

\footnotetext{
\begin{tabularx}{\dimexpr\linewidth-2em\relax}[t]{@{}l@{\hspace{0.5em}}>{\raggedright\arraybackslash}X@{}}
Code at: & \url{https://anonymous.4open.science/r/KletterMix-5F3F}\\
Dataset at: & \url{https://huggingface.co/datasets/AIML-TUDA/KletterMix}
\end{tabularx}
}


\section{Introduction}

Pretraining data is one of the primary determinants of language model quality. 
While model architecture, optimization, and alignment recipes are often the most visible components of language model development, the pretraining corpus defines much of what a model can learn before instruction tuning, domain adaptation, or annealing. It determines which facts, domains, registers, styles, and linguistic phenomena are available during the highest-volume stage of training. Recent progress in open language modeling has therefore been driven not only by scale, but also by increasingly careful pretraining mixtures~\cite{gao2020pile,soldaini-etal-2024-dolma,li2024datacomplm,diao2025climb}.\looseness=-1

This progress has been uneven across languages. English has benefited from large, diverse, and well-documented pretraining corpora, whereas German-language resources remain comparatively less mature: they are often derived from noisy web crawls, embedded as subsets of multilingual corpora, or released with limited documentation and validation through controlled training experiments~\citep{ortiz-suarez-etal-2020-oscar,kreutzer-etal-2022-quality,dodge-etal-2021-documenting,scheible-etal-2024-gottbert,dada2023on,burns-etal-2026-aleph}. This gap matters because strong German model behavior cannot be assumed to emerge from English-centric data alone. German morphology, compounding, capitalization, regional variation, and domain-specific register all shape the effectiveness of German language modeling.

One response is to collect and filter more native German web text, and recent German and European efforts have made substantial progress in that direction~\cite{burns-etal-2026-aleph,pfister-etal-2025-llammlein,gienapp2025germancommons,ali2024teuken}. However, native crawling alone does not automatically reproduce the source diversity, mixture design, or documentation standards of the strongest English pretraining datasets. High-quality German material is distributed across heterogeneous sources, and aggressive filtering can remove useful long-tail content while still leaving substantial noise. An alternative is to transfer the structure of a strong English pretraining mixture into German through high-quality machine translation. This can preserve topical diversity, source balance, and mixture-level design decisions, but it also introduces risks, e.g. translationese, semantic drift, source-language bias, length pathologies, and failures on long or specialized documents. Translated pretraining data must therefore be treated as a dataset-construction problem rather than as a simple data augmentation step.\looseness=-1

We introduce \datasetname{}, a \klettermixtokens{}-token German pretraining and annealing corpus built by translating ClimbMix, a recent high-quality English pretraining mixture~\cite{diao2025climb}, into German. \datasetname{} is designed to transfer the coverage and mixture structure of its English source while preserving document boundaries, document identifiers, metadata, source composition, and topical diversity. To assess translation quality at scale, we use COMETKiwi~\cite{rei-etal-2022-cometkiwi} as a reference-free quality-estimation signal and train a proxy model for corpus-level quality estimation. We combine these diagnostics with analyses of document length, source composition, topic coverage, and metadata preservation, and evaluate downstream utility in matched 0.6B-parameter pretraining and corpus-choice annealing experiments as well as a controlled 7B-parameter language-mixture annealing study.

Our contributions are as follows:
\begin{itemize}
    \item We introduce \datasetname{}, a \klettermixtokens{}-token German pretraining corpus constructed from a high-quality English source mixture while preserving document-level structure and metadata.
    \item We describe a scalable translation pipeline for pretraining data, including length-aware batching, chunking, dynamic output budgeting, and corpus-level validity checks.
    \item We document the translated corpus through COMETKiwi-based and proxy-based quality diagnostics, together with analyses of length distributions, topic coverage, source composition, and metadata preservation.
    \item We evaluate \datasetname{} at two model scales: through matched 0.6B-parameter pretraining, proxy-filtering, and corpus-choice annealing experiments, and through a 7B-parameter language-mixture annealing sweep that measures German gains alongside English retention.
\end{itemize}

Overall, \datasetname{} studies a practical route toward stronger non-English data: rather than relying exclusively on additional native web crawling, it asks whether the curation decisions, source diversity, and mixture structure of a strong English corpus can be transferred to German through careful translation, documentation, and empirical validation. Across two model scales and complementary training regimes, our results show that this approach can meaningfully strengthen German pretraining data while also highlighting the quality-control requirements that translated corpora demand.

\section{Related Work}
\paragraph{Large-scale pretraining corpora.}
Large-scale pretraining corpora have increasingly shifted from raw web-scale collection toward documented, filtered, and evaluation-driven dataset artifacts.
Early influential resources such as The Pile~\cite{gao2020pile} and ROOTS~\cite{laurencon2022roots} demonstrated the importance of diverse source mixtures, corpus documentation, and multilingual coverage for LLM pretraining.
More recent English-focused corpora such as Dolma~\cite{soldaini-etal-2024-dolma}, DCLM~\cite{li2024datacomplm}, and ClimbMix~\cite{diao2025climb} further emphasize reproducible curation pipelines, filtering, deduplication, mixture design, and downstream evaluation as central components of dataset design.
In parallel, multilingual web and open-data corpora such as mC4~\cite{xue-etal-2021-mt5}, OSCAR~\cite{ortiz-suarez-etal-2020-oscar}, CulturaX~\cite{nguyen-etal-2024-culturax}, HPLT~\cite{de-gibert-etal-2024-new,oepen2025hplt}, FineWeb2~\cite{penedo2025fineweb2}, and Common Corpus~\cite{langlais2026common} have expanded coverage beyond English, often by applying language identification, deduplication, quality filtering, provenance tracking, and license documentation at scale.
These efforts show that pretraining data quality depends not only on scale but also on source composition, filtering choices, documentation, and validation through model training.
\datasetname{} complements these corpora: rather than constructing German pretraining data solely through language-specific web crawling and filtering, we ask if the diversity and mixture decisions of a strong English pretraining corpus can be transferred to German through high-quality translation and validated through controlled training experiments.\looseness =-1

\paragraph{German and European language-model data.}
Recent German and European LLM efforts highlight both the demand for language-specific data and the difficulty of obtaining it at sufficient scale and quality.
GermanWeb~\cite{burns-etal-2026-aleph} constructs a large German pretraining corpus from Common Crawl, FineWeb2, and synthetic data using heuristic and model-based curation, and validates the resulting data through from-scratch pretraining.
LLäMmlein~\cite{pfister-etal-2025-llammlein} instead emphasizes transparency for compact German-only models, releasing German decoder models, training data, code, and checkpoints.
German Commons~\cite{gienapp2025germancommons} focuses on a different axis: verifiable licensing and provenance, assembling a large corpus of openly licensed German text across legal, scientific, cultural, political, news, economic, and web domains.
Other efforts adapt or build open models for German and European language coverage, including LeoLM~\cite{Pluester2023leolm}, Occiglot~\cite{avramidis-etal-2024-occiglot}, OpenGPT-X/Teuken~\cite{ali2024teuken}, and EuroLLM~\cite{martins2024eurollm}.
These works are complementary to \datasetname{}: they primarily improve German or European modeling through native web curation, continued pretraining, multilingual mixtures, or licensing-driven corpus construction, whereas \datasetname{} studies whether a high-quality English mixture can be transferred into German while preserving document boundaries, source metadata, and topical structure.\looseness =-1

\paragraph{Machine-translated data for pretraining.}
Machine-translated corpora provide a direct way to transfer high-resource English data into languages whose native pretraining corpora are smaller, less curated, or less diverse.
Recent work shows that this strategy can be effective when applied carefully.
Translation-based pretraining translates web-crawled documents into low-resource target languages and filters the resulting synthetic text, finding that filtered translations can be competitive with native, clean data for smaller language models~\cite{doshi-etal-2024-pretraining}.
TransWebEdu~\cite{wang-etal-2025-multilingual-language} scales this idea to LLM pretraining by translating FineWeb-Edu into multiple languages and training a multilingual model from scratch on the resulting corpus.
At the same time, translated data can introduce translationese artifacts, source-language bias, unnatural target-language style, and quality failures.
Large-scale web corpora may already contain substantial amounts of low-quality machine-translated content~\cite{thompson-etal-2024-shocking}, and German-focused curation work has cautioned that English-to-German machine translation should be accompanied by rigorous quality and naturalness checks~\cite{burns-etal-2026-aleph}.
\datasetname{} therefore treats translation not merely as data augmentation but as a dataset-construction problem: we preserve aligned document structure and metadata, measure translation quality with reference-free QE, and evaluate the translated corpus through controlled pretraining and annealing ablations.

\paragraph{Dataset documentation and translation quality assessment.}
Dataset documentation frameworks such as Datasheets for Datasets~\cite{gebru2021datasheets} and Data Statements for NLP~\cite{bender-friedman-2018-data} motivate explicit reporting of a dataset's motivation, composition, collection and processing pipeline, intended uses, limitations, and sociotechnical context.
These concerns are especially important for translated pretraining corpora, where source provenance, transformation steps, language variety, and translation artifacts all affect downstream use.
For translation quality, reference-free quality estimation models such as COMETKiwi~\cite{rei-etal-2022-cometkiwi} are attractive because they can score source--translation pairs without human references while supporting scalable corpus-level diagnostics.
\datasetname{} follows this line of work by coupling corpus-level documentation with translation-quality analyses, metadata preservation checks, and model-training validation; we treat automatic QE as a scalable diagnostic rather than a replacement for downstream evaluation or manual inspection.

\section{\datasetname{} Pipeline}
\label{sec:method}
\label{sec:dataset-variants}

\begin{figure}[t]
    \centering
    \resizebox{\linewidth}{!}{\input{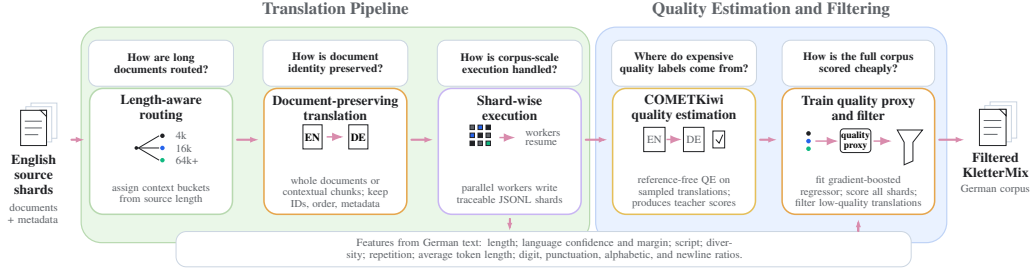}}
    \caption{%
    Overview of the \datasetname{} pipeline. English source shards are routed into length-aware context buckets, translated with a document-preserving procedure that keeps document identifiers, order, and metadata, and executed shard-wise with parallel workers. A stratified sample of translations is scored with COMETKiwi to obtain reference-free quality labels, which supervise a cheap gradient-boosted proxy that scores the full corpus from target-side features and filters low-quality translations.
    }
    \label{fig:KletterMix_pipeline}
\end{figure}

We construct \datasetname{} by translating a high-quality English pretraining mixture into German while preserving document boundaries, source metadata, and mixture structure. The pipeline consists of five stages: source-record normalization, length-aware routing, document-preserving translation, scalable shard-wise execution, and post-hoc quality estimation. This section describes the construction procedure; dataset variants and corpus-level quality analyses are reported in \autoref{sec:dataset-variants} and \autoref{sec:translation-insights}, with implementation and proxy-validation details in \appref{app:pipeline-implementation}.

\textbf{Source corpus and record structure}\quad
The source corpus is stored as sharded JSONL files. Each record contains a document identifier, the English source text, and metadata inherited from the original mixture, including source cluster, source location, and approximate source length. We preserve this information throughout the translation pipeline. Each translated record, therefore, consists of the original document identifier, the German translation, the unchanged metadata fields, and additional processing metadata such as context bucket, chunking status, translation configuration, and quality-estimation outputs.
Preserving document-level structure is important for two reasons. First, it allows \datasetname{} to retain the source mixture design of the English corpus rather than collapsing the data into sentence-level translation pairs. Second, it enables direct corpus-level comparisons between the English source and German target, including analyses of length ratios, source composition, cluster-level quality, and metadata preservation.\looseness=-1

\textbf{Length-aware routing}\quad
Pretraining corpora contain documents with highly variable lengths, from short snippets to long web pages and technical documents. A single translation configuration is inefficient for such data: short documents waste context budget, while long documents may exceed the effective context or generation limits of the translation model. We therefore route documents into length-aware buckets before translation.
Concretely, each document is assigned to a bucket according to its approximate source length. We use a small set of context buckets chosen to cover short, medium, long, and overflow documents under the practical context and generation limits of the translation setup. The same document-preserving procedure is used for all buckets, with bucket-specific source and target budgets. This reduces padding and scheduling inefficiency while allowing longer documents to be handled without imposing the most restrictive configuration on the entire corpus.\looseness =-1

\textbf{Document-preserving translation}\quad
Each source document is translated either in a single pass or through contextualized chunking. Documents that fit within the effective source budget of their assigned bucket are translated directly. Longer documents are first segmented into sentences and then greedily packed into source chunks up to a fixed token budget. If an individual sentence exceeds the chunk budget, we split it at the token level as a fallback. This ensures that every source document can be translated while preserving the original document identity.

For chunked documents, chunks are translated sequentially. To improve discourse continuity, terminology consistency, and local coherence across chunk boundaries, the prompt for chunk $t$ includes a truncated window of the German translation produced for chunk $t-1$. The model is instructed to use this previous translation only as context and to output only the translation of the current source chunk. The final German document is obtained by concatenating translated chunks in their original order.
This design makes a trade-off between scalability and document coherence. Full-document translation is preferred whenever possible, but contextualized chunking allows us to translate documents that exceed the practical context window while still exposing the model to local context from the preceding chunk.\looseness=-1

\textbf{Dynamic target-side budgeting}\quad
English-to-German translation can change document length depending on genre, domain, and formatting. We therefore do not use a fixed generation cap for all documents. Instead, the maximum target-side generation length is derived from the source chunk length and capped by a global maximum. Formally, for a source chunk of length $\ell_{\mathrm{src}}$, we set the target budget to $\ell_{\mathrm{tgt}}^{\max} = \min\left( L_{\max}, \left\lceil \alpha \ell_{\mathrm{src}} + \beta \right\rceil \right),$
where $L_{\max}$ is the maximum generation budget and $\alpha,\beta$ are chosen to allow moderate target-side expansion. This avoids over-allocating decoding budget to short chunks while reducing truncation risk for long or expansion-heavy documents.

\textbf{Scalable shard-wise execution}\quad
Translation is performed shard-wise. Multiple workers process disjoint subsets of the source corpus and write translated records incrementally. Workers preserve the original shard and document identifiers, which makes translated outputs traceable to the source corpus and simplifies resumption after interruption.

To make the pipeline robust to transient failures, workers write intermediate attempt files and finalize a shard only after all records in that shard have been processed. When a run is interrupted, processing resumes from the last completed record rather than restarting the shard. Translation requests are distributed across a pool of model endpoints. Workers monitor endpoint availability and requeue unfinished documents when an endpoint becomes unavailable. These engineering choices do not affect the dataset definition, but they are important for reproducibility at corpus scale.
We motivate the selected translation model, precision, and speculative-decoding configuration in \appref{app:translation-model-selection} and give the concrete translation, prompt, and serving settings in \appref{sec:translation-pipeline-configs}.

\textbf{Pilot subset for translation-quality estimation}\quad
Full-corpus reference-free quality estimation is computationally expensive. We therefore first construct a pilot subset for quality analysis and proxy-model training. The pilot subset is sampled in a stratified manner over source clusters. Within each cluster, documents are ranked by a stable hash of their document identifier, and the top-ranked documents are selected until the cluster-specific quota is reached. This produces a reproducible sample while preserving the mixture structure of the source corpus.
The pilot subset is translated with the same pipeline as the full corpus. We then score the resulting source--translation pairs with COMETKiwi~\cite{rei-etal-2022-cometkiwi}, a reference-free quality-estimation model. We use these scores as corpus-level diagnostic signals rather than as a substitute for downstream training evaluation. In particular, COMETKiwi is used to identify quality variation across length buckets, source clusters, and document types, and to supervise a cheaper proxy model for full-corpus scoring.

\textbf{Proxy-based quality annotation}\quad
Full-corpus COMETKiwi scoring is too expensive to use as the only annotation mechanism for every translated document. We therefore train a cheap proxy annotator to predict COMETKiwi scores from the COMETKiwi-scored pilot data described above. The deployed proxy is a gradient-boosted regression model with a target-only feature set: it reads only the translated German document and inexpensive metadata derived from that text. This deployment choice is central to the system design. Because the proxy does not need the original English source document, it can score released German translated shards directly and avoids a costly source-rehydration pass over the full corpus.\looseness=-1

After translation, we run GlotLID~\cite{kargaran-etal-2023-glotlid} on each German target document to obtain language-identification signals for the proxy. GlotLID is loaded as a fastText model and applied to the translated text after replacing newlines with spaces. We restrict scoring to a fixed set of common language/script labels and compute normalized probabilities over that set. From this pass, we derive whether the top label is German Latin (\texttt{deu\_Latn}), the normalized German-Latin probability, a clipped logit transform of that probability, the top-1/top-2 probability margin, and the script extracted from the top predicted label. These GlotLID-derived signals are combined with cheap target-side text-shape features: length, lexical diversity, token repetition, average token length, and character-composition ratios.\looseness=-1

\autoref{tab:proxy-features} summarizes the deployed feature set. The GlotLID features capture wrong-language output and low-confidence German predictions; script, length, and character-composition features capture abnormal text shape, formatting artifacts, and suspicious character mixtures; and lexical-diversity and repetition features capture degenerate or repetitive generations. The proxy cannot directly measure semantic adequacy, because it does not see the English source. Its role is therefore not to replace source-aware evaluation or human inspection, but to provide a scalable corpus-level quality signal that is validated against COMETKiwi and targets practical failure modes that matter at release scale.

We validate the proxy on a disjoint 18{,}275-document split scored with COMETKiwi; full agreement metrics are reported in \autoref{tab:proxy-validation}. The target-only proxy shows strong agreement with COMETKiwi and low absolute error, making it suitable as a scalable ranking and filtering signal rather than as a replacement for source-aware quality evaluation. During development, source-aware variants did not improve validation agreement enough to justify rehydrating the English source text for full-corpus scoring. We therefore deploy the target-only model as the full-corpus annotator.

\autoref{fig:sub_cluster_proxy_score_quantiles} uses the proxy scores as a cluster-level diagnostic over the inherited source clusters. For each fixed source-cluster identifier, we compare the 12B-token training subset and the full translated corpus by plotting the 10th--90th percentile interval, the 25th--75th percentile interval, and the median target-only proxy score. The panel is not a thresholding plot; its purpose is to show whether proxy-estimated translation quality is broadly consistent across source clusters and whether the fixed-budget subset preserves the full-corpus quality profile.

\begin{figure}[t]
    \centering
    \includegraphics[width=0.75\linewidth]{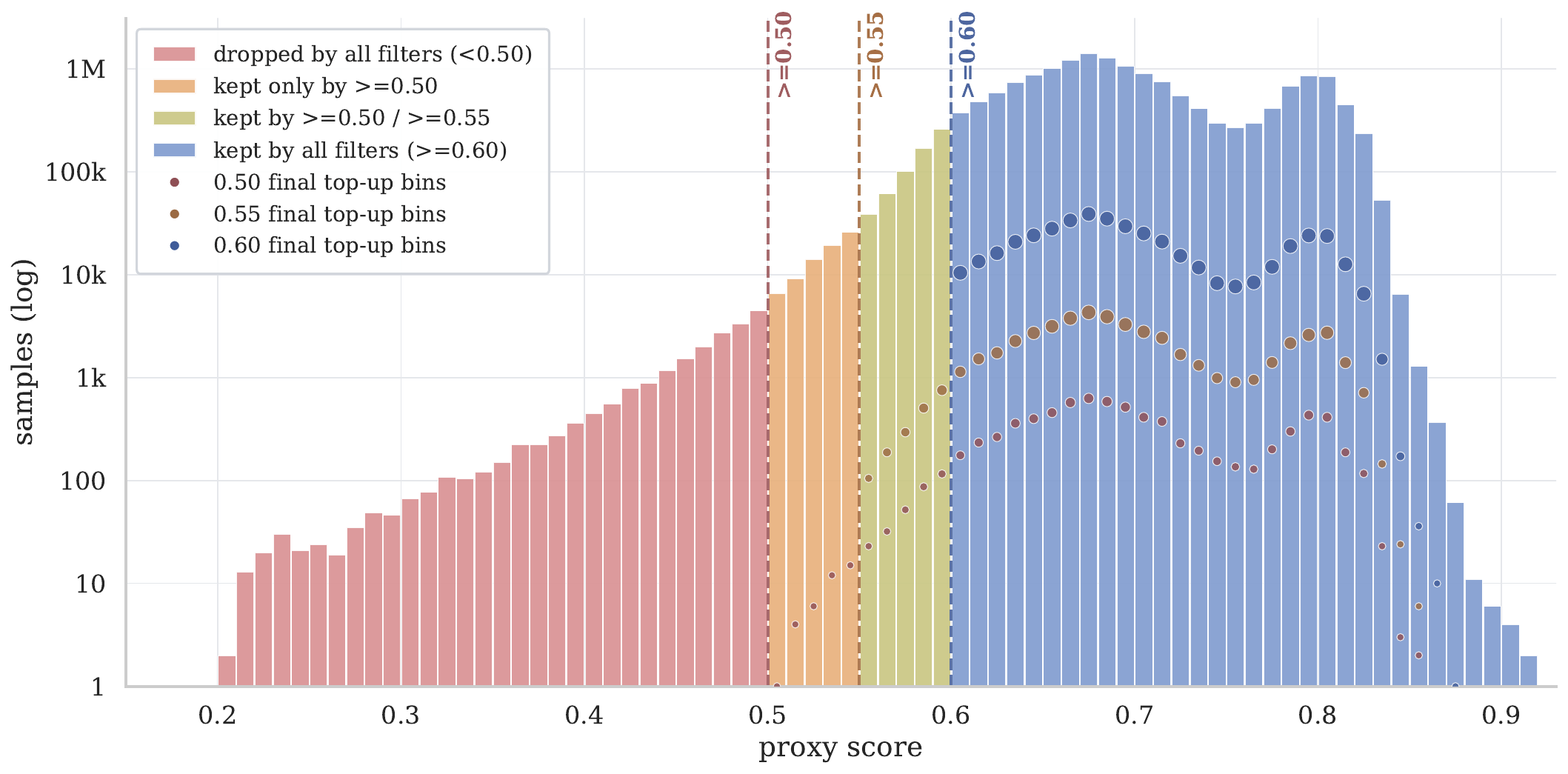}
    \caption{Proxy-score distribution and filtering thresholds used to construct the three 12B-token filtering ablations. The histogram shows the full translated-corpus proxy-score distribution on a log-count axis. Dashed vertical lines mark the retained-document thresholds at 0.50, 0.55, and 0.60; colored regions indicate which documents are removed or retained by progressively stricter filters.}
    \label{fig:proxy_score_thresholds}
\end{figure}

\textbf{Proxy-filtered dataset variants}
The primary release of \datasetname{} is the unfiltered translated corpus, excluding only records that fail basic validity checks such as empty output, missing metadata, or severe language-identification failure. In addition, we construct three proxy-filtered 12B-token training splits by retaining only translated documents whose predicted COMETKiwi proxy score exceeds a fixed threshold:
$\hat{q}_{\mathrm{proxy}} \geq 0.50$, $\hat{q}_{\mathrm{proxy}} \geq 0.55$ and $\hat{q}_{\mathrm{proxy}} \geq 0.60$.
These thresholds are not intended as universal quality cutoffs. They define filtering ablations: under a fixed 12B-token training budget, we ask whether removing progressively more low-scoring translated text improves optimization behavior, held-out validation loss, or downstream benchmark performance. Unless otherwise stated, corpus-level analyses refer to the unfiltered translated corpus, while training experiments compare two external German-data baselines, FineWeb2-DE and GermanWeb, against unfiltered \datasetname{} and the three threshold-filtered \datasetname{} splits. \autoref{fig:proxy_score_thresholds} visualizes the full proxy-score distribution and marks the three retained-document thresholds used for these ablations.

Because \datasetname{} is constructed by translating an existing English mixture while preserving document boundaries and metadata, the resulting German corpus can be characterized along the same structural axes as its source. We therefore describe the release in terms of document-length distributions, length-bucket consistency, and cluster-level proxy-score patterns across both the 12B-token subset and the full corpus. These corpus-level views serve as documentation and quality-control signals: they show whether the translated data preserves the expected length profile of a high-quality pretraining mixture, reveal unusually short translations in long-document buckets, and highlight source clusters that may warrant closer audit. The proxy thresholds above define the optional filtered variants, while the analyses below characterize the structure and quality profile of the released corpus. \appref{app:translation-insights-details} gives the cluster-labeling procedure and qualitative examples used for audit.

\section{Translation Insights}
\label{sec:translation-insights}
\begin{figure}[ht]
    \centering
    \begin{subfigure}[t]{0.48\linewidth}
        \centering
        \includegraphics[width=\linewidth]{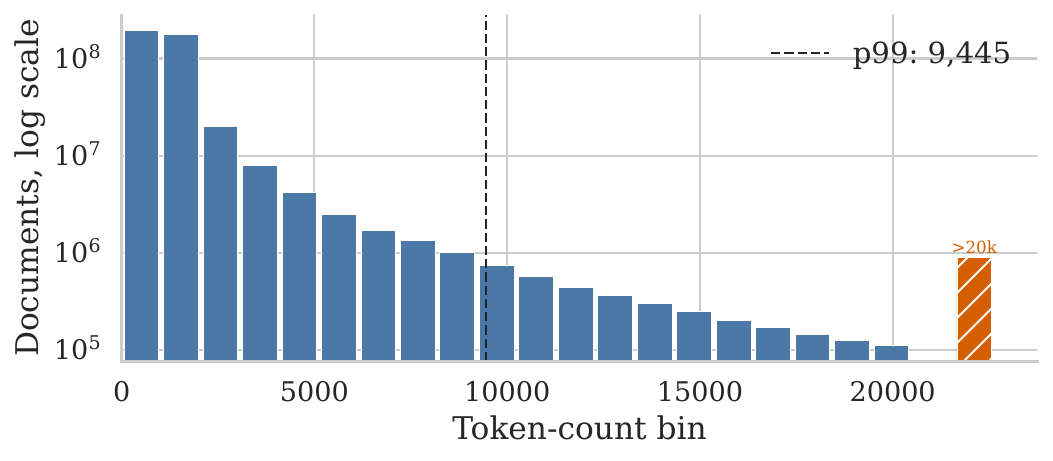}
        \caption{Translated document lengths. The corpus has a heavy-tailed profile, with most documents below 10k tokens and a long tail beyond 20k tokens.}
        \label{fig:sub_token_count_histogram}
    \end{subfigure}
    \hfill
    \begin{subfigure}[t]{0.48\linewidth}
        \centering
        \includegraphics[width=\linewidth]{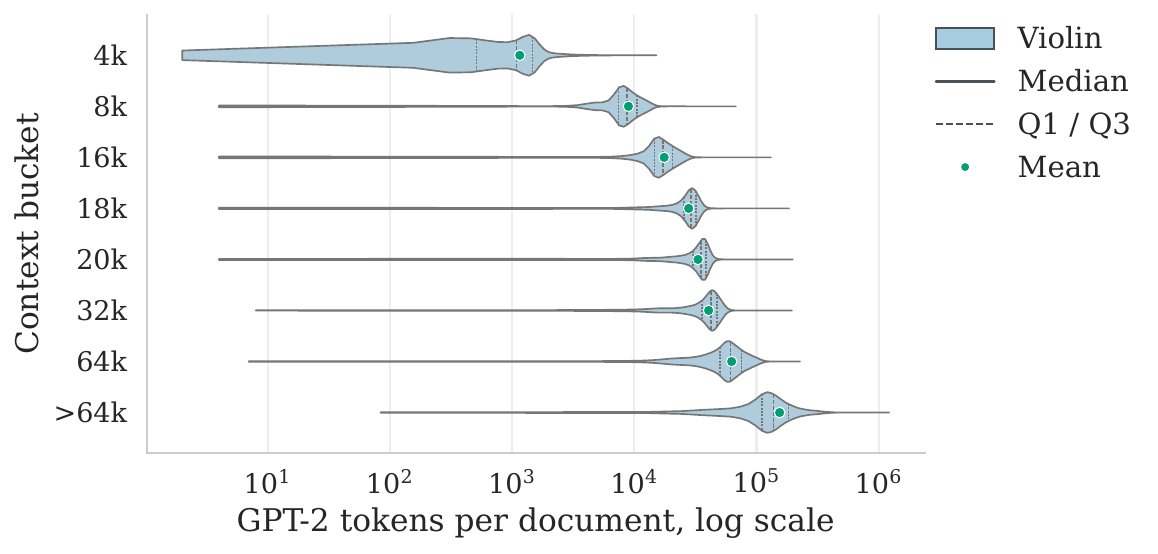}
        \caption{Target lengths by source-length bucket. Long lower tails highlight unusually short translations in long-context buckets.}
        \label{fig:sub_bucket_violin}
    \end{subfigure}
    \vskip\baselineskip
    \begin{subfigure}[t]{0.7\linewidth}
        \centering
        \includegraphics[width=\linewidth]{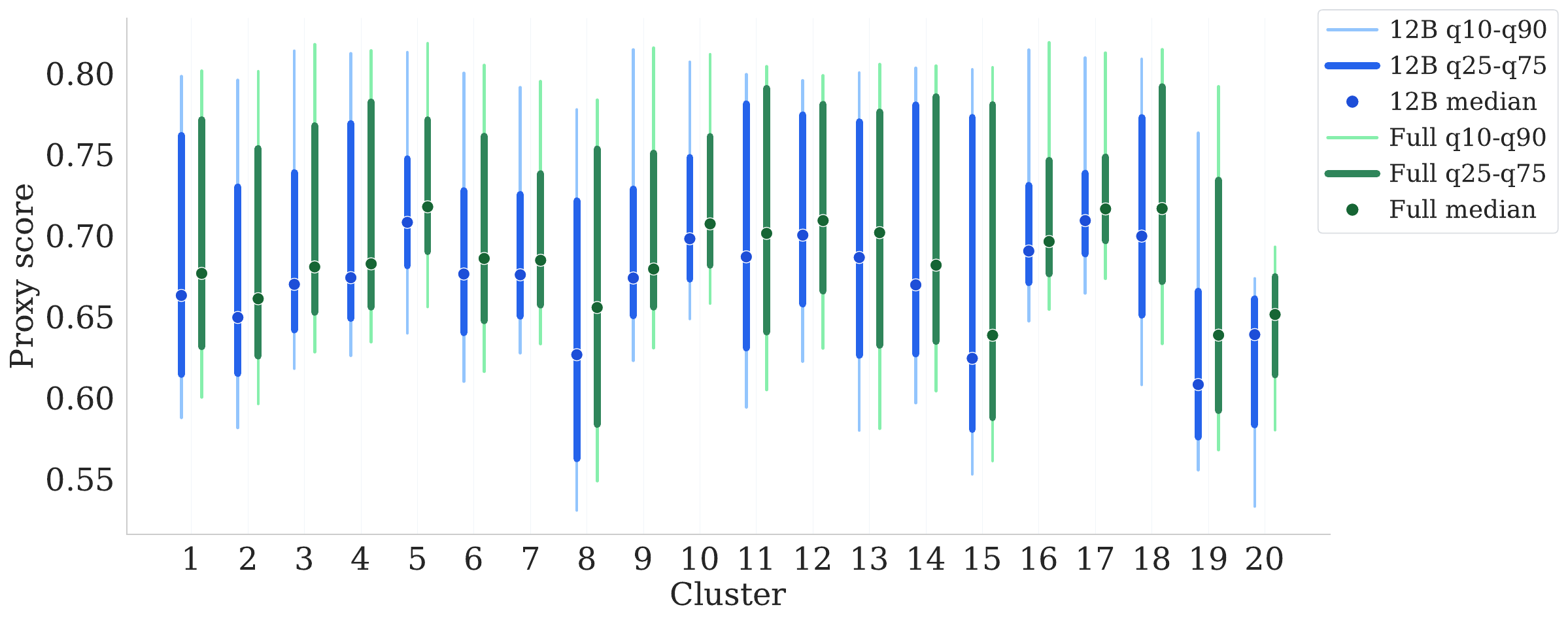}
        \caption{Proxy-score quantiles by inherited source cluster for the 12B-token subset and the full corpus. Intervals show 10th--90th and 25th--75th percentiles; points show medians.}
        \label{fig:sub_cluster_proxy_score_quantiles}
    \end{subfigure}
    \caption{
    Corpus diagnostics for full \datasetname{} and the 12B subset.
    The length plots summarize global target-document lengths and bucket-level consistency, while the proxy-score comparison checks whether fixed-budget sampling preserves the quality profile across inherited source clusters.\looseness=-1}
    \label{fig:translation_insights}
\end{figure}

\autoref{fig:translation_insights} summarizes these corpus-level diagnostics for the full \datasetname{} release and the 12B-token training subset. The two length-based panels inspect global target-document lengths and bucket-level length consistency, while the cluster-level proxy-score panel checks whether fixed-budget sampling preserves the full-corpus quality profile across inherited source clusters.

\autoref{fig:sub_token_count_histogram} shows the overall document-token distribution. The corpus follows a heavy-tailed length profile: most translated documents are relatively short, while a smaller number of documents extend into the long-context regime. This structure is typical of heterogeneous web and pretraining mixtures and motivates the length-aware translation pipeline described in \autoref{sec:method}.

\autoref{fig:sub_bucket_violin} provides a more fine-grained view by comparing translated document lengths within each canonical source-length bucket. Because token counts are tokenizer-dependent and can vary across languages, we do not expect a fixed one-to-one correspondence between English source length and German target length. Nevertheless, English-to-German translation often leads to surface-length expansion, and German may require more subword tokens under a given tokenizer due to language-specific morphology and tokenizer fertility effects~\cite{ahia2023all,rust-etal-2021-good}. Thus, documents assigned to longer source buckets should generally also produce relatively long German translations, up to expected variation from tokenization, genre, and translation style.

The violin plot shows that this is usually the case, but it also reveals long lower tails: some documents in long-context buckets yield translated outputs much shorter than expected. These cases are suspicious because the source document was long enough to require a larger context bucket, yet the resulting German document is comparatively short, which may indicate truncation, dropped content, or other translation failures. We treat these lower tails as corpus diagnostics and audit signals rather than as the definition of the filtered training splits; the current filtering ablations use only the target-only proxy score thresholds described in the proxy-filtered variants introduced in \autoref{sec:method}.

Finally, \autoref{fig:sub_cluster_proxy_score_quantiles} compares cluster-level proxy-score quantiles for the 12B-token training subset and the full translated release. Across clusters, the panel shows the 10th--90th percentile range, the 25th--75th percentile range, and the median target-only proxy score, making it a corpus diagnostic rather than a filtering-threshold figure. The comparison checks whether the fixed-budget subset used in training ablations preserves the release's cluster-level quality profile and highlights clusters whose proxy-score distributions warrant closer audit.

Overall, these analyses suggest that \datasetname{} preserves a diverse and heavy-tailed pretraining mixture while also exposing document-level and cluster-level quality-control signals. In particular, unusually short translations for long-bucket source documents and cluster-specific proxy-score variation motivate continued corpus auditing, while the controlled filtering experiments in \autoref{sec:training-ablations} isolate the effect of target-only proxy-score thresholds.

\section{Training Ablations}
\label{sec:training-ablations}

We evaluate whether \datasetname{} improves model training relative to established German pretraining data. While the previous sections analyze the translated corpus directly, this section tests its usefulness under controlled model, compute, and token-budget conditions. We first compare German corpora through matched 0.6B-parameter pretraining and corpus-choice annealing, and then vary the \datasetname{} fraction in a 7B-parameter annealing mixture.

\textbf{Matched 0.6B setup.}\quad
We conduct controlled pretraining ablations using Qwen3-0.6B~\cite{yang2025qwen3} as the base architecture and follow a Megatron-LM training recipe~\cite{megatron-lm}. Each run is trained on a matched German 12B-token subset, corresponding to $\sim$20 tokens per parameter for a 0.6B-parameter model, following the Chinchilla scaling recommendation~\cite{hoffmann2022training}. We compare \datasetname{} against FineWeb2-DE~\cite{penedo2025fineweb2} and GermanWeb~\cite{burns-etal-2026-aleph} under matched conditions, including model architecture, optimizer, batch size, learning-rate schedule, tokenizer, token budget, and validation protocol (\appref{sec:pretraining-setup}).\looseness=-1

For each corpus, we construct the training subset with a deterministic, token-budgeted stratified sampler. The sampler estimates document-level token counts, preserves the main structure of each source corpus through corpus-specific strata, and selects documents by hashing stable document identifiers until the target token budget is reached. For \datasetname{}, the strata follow the inherited source-cluster and context-length structure; for FineWeb2-DE and GermanWeb, they follow the available source and crawl metadata. Validation is performed on separate held-out splits produced with the same sampling procedure but disjoint selection windows, ensuring that validation documents are not included in the corresponding training subsets. We track training loss throughout optimization and evaluate validation loss and perplexity at regular intervals. In addition to the unfiltered version of \datasetname{}, we evaluate proxy-filtered variants constructed using the quality model described in \autoref{sec:method}.\looseness =-1

\begin{figure}[t]
    \centering
    \begin{subfigure}[b]{0.48\linewidth}
        \centering
        \includegraphics[width=\linewidth]{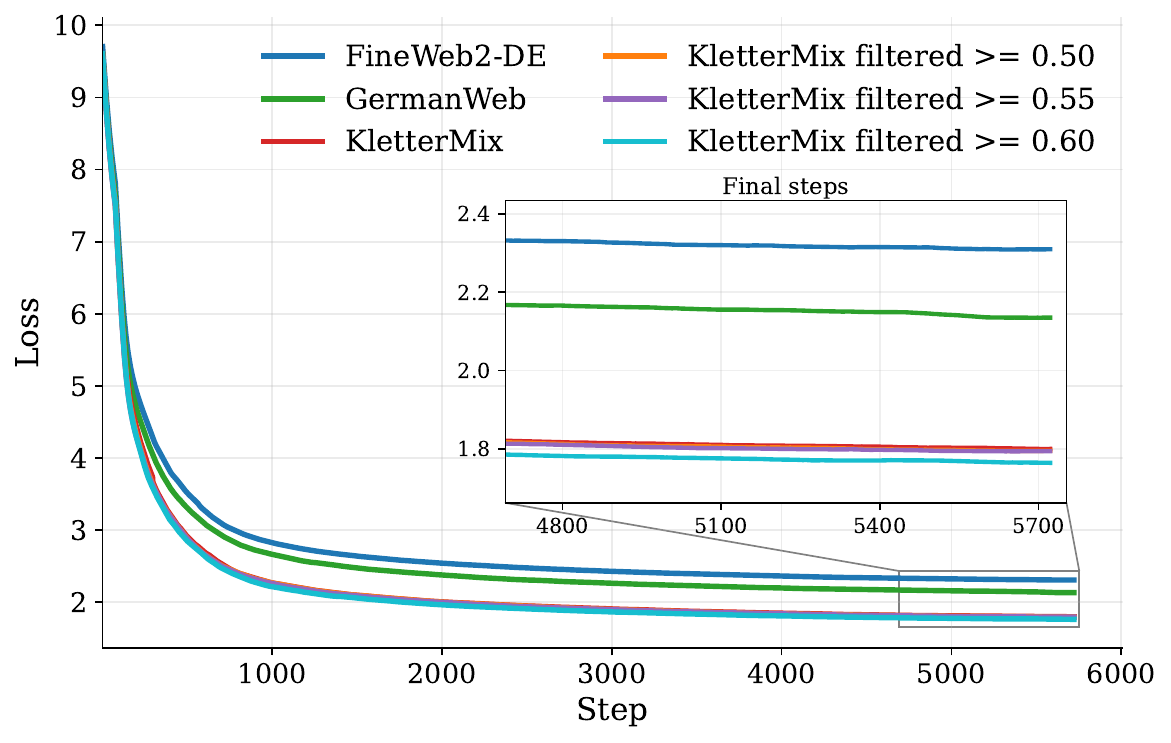}
        \caption{Pretraining dynamics for Qwen3-0.6B trained on matched 12B-token German subsets. 
        \datasetname{} exhibits lower training loss under the matched recipe.}
        \label{fig:training_loss}
    \end{subfigure}
    \hfill
    \begin{subfigure}[b]{0.48\linewidth}
        \centering
        \includegraphics[width=\linewidth]{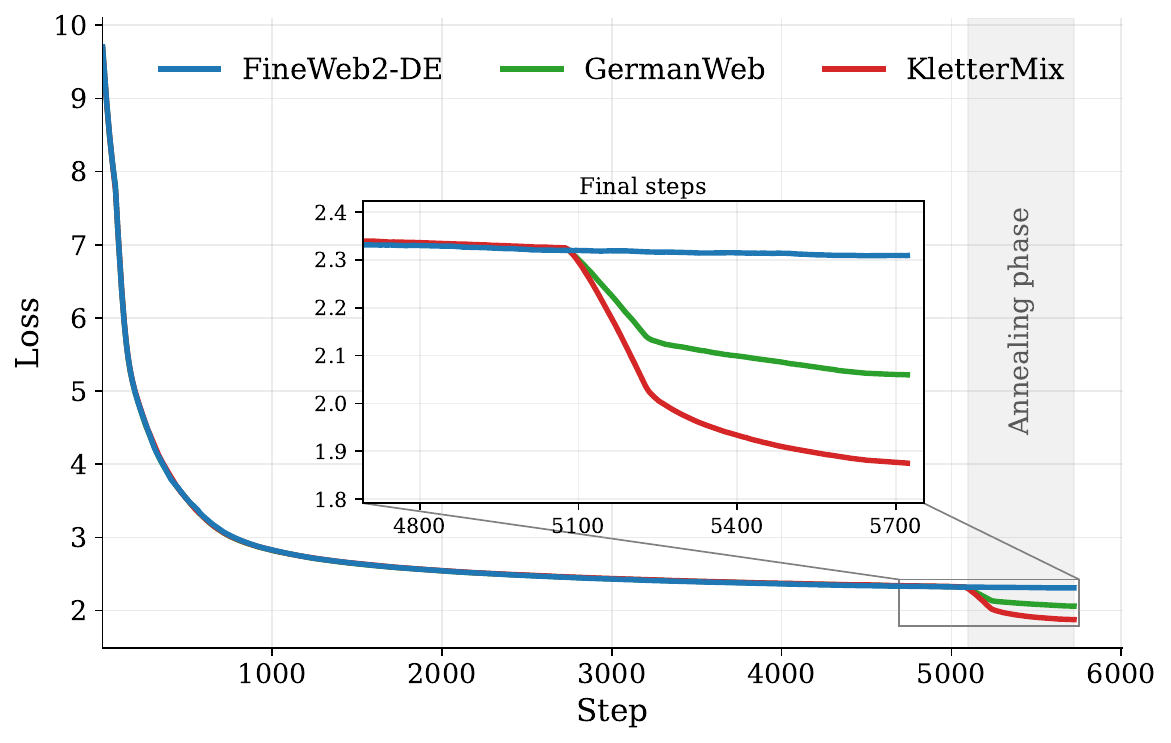}
        \caption{
        Annealing dynamics for Qwen3-0.6B trained for 5100 steps on FineWeb2-DE and then annealed on GermanWeb or \datasetname{}.\looseness=-1
        }\label{fig:annealing_loss}
    \end{subfigure}
    \caption{
     Training and annealing dynamics on matched 12B-token German subsets. Across both training regimes, KletterMix attains lower loss than existing German corpora, suggesting more favorable optimization dynamics. Benchmark performance remains the primary criterion for evaluating utility. \looseness=-1
    }
    \label{fig:training_dynamics}
\end{figure}

\textbf{0.6B training dynamics.}\quad
\autoref{fig:training_loss} shows training for models trained on 12B-token subsets. Across the full training run, \datasetname{} reaches lower training loss than FineWeb2-DE and GermanWeb under the same optimization setup. More importantly, the advantage also appears on the held-out validation set (see \autoref{fig:app_validation_loss_klettermix_zoom}): the \datasetname{} model maintains consistently lower validation loss throughout training. This suggests that the translated corpus is not merely easier to fit, but provides a training signal that transfers to held-out German text.
The validation gap is visible early in training and persists through the final checkpoints. This behavior indicates that \datasetname{} improves sample efficiency under a fixed token budget. We report validation perplexity and accuracy in \appref{app:training-results}.

\begin{table}[t]
\centering
\caption{Downstream 5-shot accuracy on German evaluations under matched 12B-token training conditions. Bold and underline mark the best and second-best point estimates within each metric column across the full table; ties are marked identically. Rows under FineWeb2-DE annealing continue from the same FineWeb2-DE checkpoint. Details on score calculations are in \appref{app:notes_benchmark_results}}
\label{tab:benchmark_results}
\small
\setlength{\tabcolsep}{4.0pt}
\renewcommand{\arraystretch}{0.95}

\newcommand{\evalscore}[2]{\ensuremath{#1{\scriptscriptstyle\,\pm\,#2}}}
\newcommand{\bestevalscore}[2]{\ensuremath{\mathbf{#1{\scriptscriptstyle\,\pm\,#2}}}}
\newcommand{\secondevalscore}[2]{\underline{\evalscore{#1}{#2}}}

\begin{tabular}{@{}lccccc@{}}
    \toprule
    Run & MMLU & PIQA & HellaSwag & ARC-C & Core Avg. \\
    \midrule

    \rowcolor{gray!20}\multicolumn{5}{@{}l}{\textit{Independent pretraining}}&\phantom{123}\\[-0.1em]
    \rowcolor{gray!20}GermanWeb
        & \bestevalscore{30.0}{2.3}
        & \evalscore{63.0}{4.9}
        & \evalscore{31.2}{0.5}
        & \evalscore{23.1}{1.2}
        & \evalscore{36.8}{1.4} \\
    \rowcolor{gray!20}FineWeb2-DE
        & \evalscore{28.7}{2.3}
        & \bestevalscore{70.0}{4.6}
        & \evalscore{31.5}{0.5}
        & \evalscore{23.0}{1.2}
        & \evalscore{38.3}{1.3} \\

    \addlinespace[0.35em]
    \rowcolor{gray!0}\multicolumn{5}{@{}l}{\textit{FineWeb2-DE annealing}}&\phantom{123}\\[-0.1em]
    \rowcolor{gray!0}\hspace{0.7em}$\rightarrow$ GermanWeb
        & \bestevalscore{30.0}{2.3}
        & \evalscore{65.0}{4.8}
        & \evalscore{32.2}{0.5}
        & \evalscore{23.1}{1.2}
        & \evalscore{37.6}{1.4}\\
    \rowcolor{gray!0}\hspace{0.7em}$\rightarrow$ \datasetname{}
        & \secondevalscore{29.0}{2.3}
        & \secondevalscore{69.0}{4.6}
        & \evalscore{34.2}{0.5}
        & \evalscore{25.2}{1.3}
        & \secondevalscore{39.4}{1.3} \\

    \addlinespace[0.35em]
    \rowcolor{gray!20}\multicolumn{5}{@{}l}{\textit{\datasetname{} pretraining variants}}&\phantom{123}\\[-0.1em]
    \rowcolor{gray!20}\datasetname{}
        & \secondevalscore{29.0}{2.3}
        & \evalscore{65.0}{4.8}
        & \secondevalscore{34.4}{0.5}
        & \secondevalscore{26.5}{1.3}
        & \evalscore{38.7}{1.4} \\
    \rowcolor{gray!20}\datasetname{}-Filt$_{0.60}$
        & \evalscore{28.5}{2.3}
        & \bestevalscore{70.0}{4.6}
        & \bestevalscore{34.6}{0.5}
        & \bestevalscore{27.5}{1.3}
        & \bestevalscore{40.2}{1.3} \\

    \bottomrule
\end{tabular}
\end{table}

\textbf{Benchmark evaluation.}\quad\autoref{tab:benchmark_results} reports downstream benchmark performance for the matched 0.6B German training experiments. We evaluate all checkpoints with lm-eval-harness~\cite{eval-harness} using 5-shot accuracy. The German suite uses the German or multilingual task configurations corresponding to MMLU~\cite{hendrycks2021measuring,global-mmlu}, PIQA~\cite{bisk2020piqa,chang2025globalpiqaevaluatingphysical}, HellaSwag~\cite{zellers-etal-2019-hellaswag,thellmann2024multilingualllmevaluationeuropean}, and ARC-Challenge~\cite{clark2018think,thellmann2024multilingualllmevaluationeuropean}. The 7B experiment below uses the same German task configurations and additionally evaluates the corresponding original English tasks. We report evaluation-set standard errors next to each accuracy. 

We use the four-task \emph{Core Avg.} as our main German aggregate,
because the four tasks cover complementary axes of reasoning transfer rather than a single narrow skill. MMLU probes broad multitask knowledge and problem solving across academic and professional subjects~\cite{hendrycks2021measuring, global-mmlu}. PIQA targets physical commonsense reasoning, testing whether a model can choose plausible actions involving everyday objects and affordances~\cite{chang2025globalpiqaevaluatingphysical,bisk2020piqa}. HellaSwag evaluates grounded commonsense inference by asking the model to select the plausible continuation of an everyday scenario, with distractors that are fluent but semantically implausible~\cite{zellers-etal-2019-hellaswag, thellmann2024multilingualllmevaluationeuropean}. ARC-Challenge evaluates hard science-style question answering, where models must combine facts rather than only match surface patterns~\cite{clark2018think, thellmann2024multilingualllmevaluationeuropean}. Together, these tasks provide four complementary probes: broad stored knowledge, physical affordance reasoning, event-level plausibility, and science-style compositional inference.

Under this aggregate, the point estimates place every \datasetname{}-family row above the external baselines. The weakest \datasetname{} family result, unfiltered \datasetname{} at 38.7, is slightly above FineWeb2-DE at 38.3, while the validation-selected filtered split reaches the best overall point estimate of 40.2. Because PIQA is evaluated on a small 100-example subset, the Core Avg. standard errors are comparatively wide, so we interpret small aggregate gaps cautiously. The more stable signal is the recurring task-level pattern: \datasetname{} is consistently strongest on HellaSwag and ARC-C, suggesting that its main benefit lies in grounded event continuation and compositional science-style reasoning rather than uniform gains on every benchmark.\looseness=-1

\textbf{0.6B corpus-choice annealing.}\quad\autoref{fig:annealing_loss} and \autoref{tab:benchmark_results} isolate a different question from the independent pretraining ablations: once a model has already learned from FineWeb2-DE, which corpus provides the better final direction? Annealing FineWeb2-DE on GermanWeb raises MMLU to 30.0 and slightly improves HellaSwag, but it leaves ARC-C essentially unchanged and reduces PIQA, yielding a Core Avg. of 37.6. Annealing the same FineWeb2-DE checkpoint on \datasetname{} instead yields 39.4, improving over the FineWeb2-DE source checkpoint by +1.1 points and over the GermanWeb annealing branch by +1.8 points. The gain is concentrated where we expect a high-quality translated mixture to matter most: HellaSwag rises from 31.5 to 34.2, and ARC-C from 23.0 to 25.2, while PIQA remains close to the FineWeb2-DE baseline within its larger evaluation uncertainty. This makes the annealing result a sharper control than the independent pretraining comparison: it suggests that \datasetname{} is not merely a better initialization corpus, but a useful late-stage sharpening corpus for event coherence and science-style reasoning.

\textbf{7B language-mixture annealing.}\quad
To test whether the result extends beyond the 0.6B recipe, we independently anneal the OLMo~3 7B Stage~1 checkpoint~\cite{olmo2026olmo3} for a matched budget of $\sim$12B tokens. Five runs start from the same pretrained weights, initialize fresh optimizer and scheduler states, and use the same optimization schedule. The only intended intervention is the data mixture: \datasetname{} supplies 0\%, 5\%, 10\%, 15\%, or 20\% of the tokens, with ClimbMix supplying the remainder. For the German component, we begin with the previously defined proxy-filtered pool ($\hat{q}_{\mathrm{proxy}}\ge0.60$) and prioritize its highest-scoring documents; this procedure yields a proxy-score range of 0.70--0.90. We evaluate each checkpoint on the same four German tasks as above and on their English counterparts. Full model, optimization, mixture-construction, and evaluation details are given in \appref{app:olmo3-annealing}.

\begin{figure}[t]
    \centering
    \includegraphics[width=0.7\linewidth]{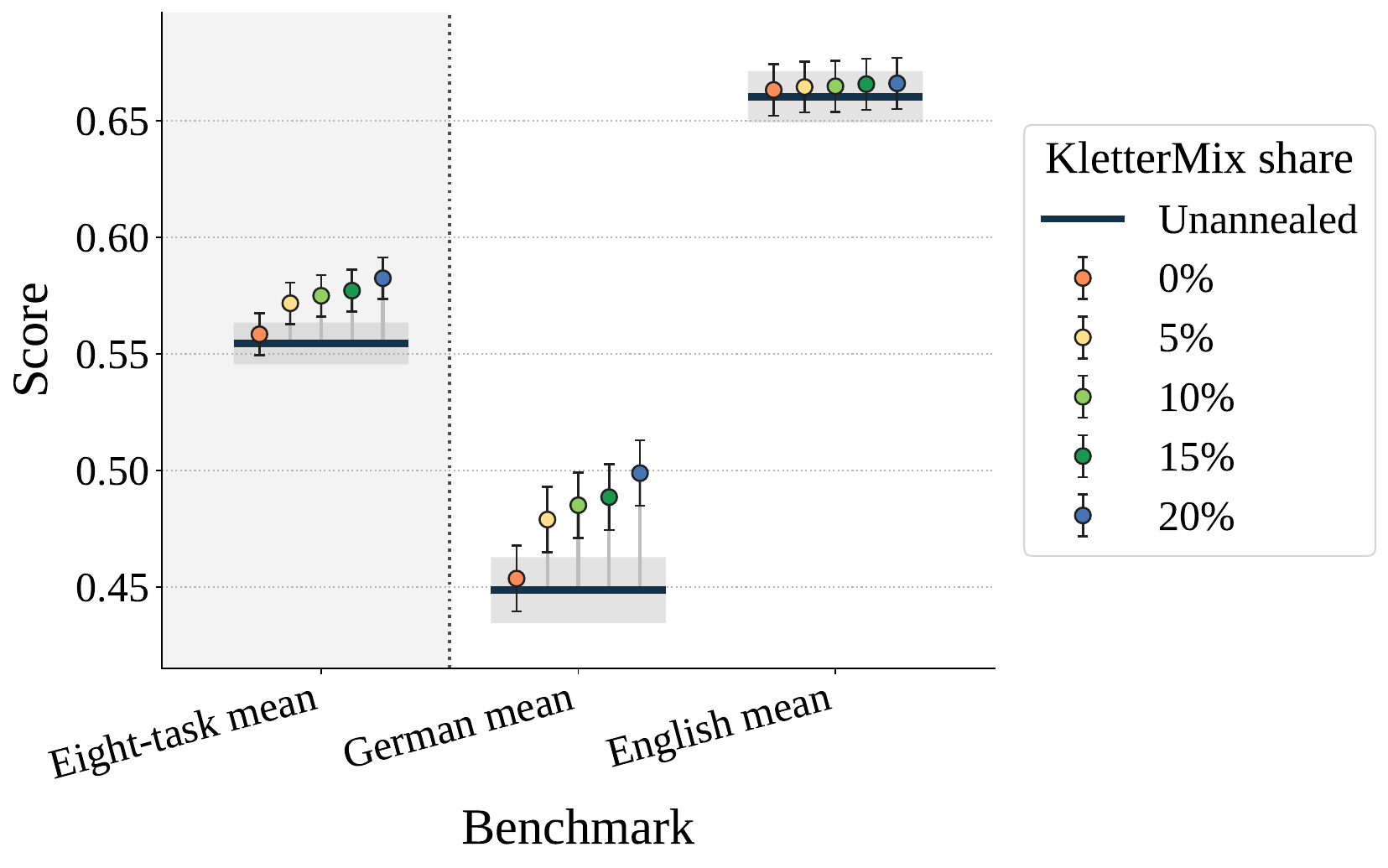}
    \caption{
Aggregate benchmark scores for the OLMo 3 7B Stage 1
checkpoint after  $\sim$12B tokens of annealing.
The dark horizontal lines and gray bands show the unannealed
reference and its standard error; colored points show
matched annealing runs containing 0\% to 20\%
KletterMix with their standard error. 
The German and English means are unweighted averages over four
tasks each, and the eight-task mean averages all eight tasks.
Error bars show standard error.
}
    \label{fig:olmo3-mixture-summary}
\end{figure}

\autoref{fig:olmo3-mixture-summary} shows a monotonic increase in the German aggregate point estimates as the \datasetname{} share grows. The matched 0\% \datasetname{} control reaches 45.4\%, compared with 47.9\%, 48.5\%, 48.9\%, and 49.9\% for the 5\%, 10\%, 15\%, and 20\% mixtures, respectively. Thus, replacing 20\% of the ClimbMix-only annealing data with \datasetname{} improves the German four-task mean by 4.5 percentage points. Over the same comparison, the English mean changes from 66.3\% to 66.6\% (+0.3 points). The unannealed Stage 1 checkpoint is shown as a separate reference;
the controlled language-mixture effect is the comparison between the
0\% KletterMix control and the 5--20\% KletterMix mixtures. 

\section{Conclusion}
\label{sec:Conclusion}

\textbf{Discussion.}\quad
Across matched 0.6B pretraining, corpus-choice annealing, and 7B
language-mixture annealing, the results support the central premise
of KletterMix: carefully translated and documented data can transfer
more than German surface form; it can transfer useful mixture structure.
The Core Avg.\ over MMLU, PIQA, HellaSwag, and ARC-C gives the
KletterMix family the strongest point estimates under our matched
recipe, while the attached evaluation-set standard errors make clear
that small aggregate gaps should not be overread. The pattern is not a uniform sweep over all tasks, and that is the point. GermanWeb remains strongest on MMLU, and FineWeb2-DE remains highly competitive on PIQA, but \datasetname{} is strongest where coherent document structure and dense explanatory content should matter most: HellaSwag and ARC-C. We therefore interpret the result as evidence for \emph{reasoning transfer through data curation}, not simply as evidence that translated text dominates native German web text.

The two annealing experiments strengthen this interpretation in complementary ways. At 0.6B parameters, continuing the same FineWeb2-DE checkpoint on \datasetname{} gives a better Core Avg. than continuing it on GermanWeb, with gains concentrated on HellaSwag and ARC-C. At 7B parameters, increasing the \datasetname{} share from 0\% to 20\% within a fixed 12B-token annealing budget yields a graded increase in the German aggregate point estimates, while the English aggregate remains nearly unchanged. The first comparison asks which German corpus is the better final steering signal; the second isolates the amount of \datasetname{} in a larger pretrained model's annealing mixture. Together, they suggest that \datasetname{} is useful not only as an initialization corpus but also as a late-stage language-adaptation component. The filtering results add a further qualification: proxy scores are useful for ranking data under a fixed budget, but not as a universal quality law, since MMLU does not improve monotonically with stricter filtering.\looseness=-1

\textbf{Limitations.}\quad
Despite these positive results, \datasetname{} has several limitations. Because the corpus is derived from a mixture of English sources, it may inherit the topical, cultural, geographic, stylistic, and licensing biases of those sources, even after translation into German. Machine translation can also introduce translationese, semantic drift, unnatural German style, inconsistent terminology, or failures in long, highly specialized documents. Our COMETKiwi-based and proxy-based quality estimates provide scalable diagnostics, but they are not substitutes for human evaluation, downstream task evaluation, or careful inspection of problematic domains. The model-based evidence  spans 0.6B and 7B parameters, however, each condition has one training run, both regimes use an $\sim$12B-token budget, the 7B experiment uses a single starting checkpoint and a quality-ranked \datasetname{} subset; future work should test larger models, additional random seeds, and broader downstream benchmark suites.\looseness=-1

\textbf{Evaluative role}. \datasetname{} is intended as both a German pretraining corpus and an aligned resource for evaluating translation-based data curation. Because each German document preserves the source document identifier, metadata, source cluster, and length bucket, the corpus supports controlled studies of: (i) whether a high-quality English mixture can be transferred to German under fixed token and compute budgets; (ii) how translation quality varies by document length, source cluster, and text type; and (iii) how proxy-based filtering affects German language-model training. \datasetname{} should not be interpreted as a replacement for native German data, as evidence that translated data is culturally representative of German-language text, or as a guarantee that proxy scores measure semantic adequacy for individual documents.

\textbf{Future Work.}\quad Future work should extend this approach along several dimensions. First, stronger filtering could target translation failures not fully captured by length or language-identification diagnostics, such as URL-only documents, boilerplate, duplicated content, or subtle semantic drift. Second, manual audits and benchmark evaluations can better characterize naturalness, factual preservation, and downstream utility across domains. Third, the same document-preserving translation pipeline could be applied to other languages such as French, Italian, and Spanish, enabling systematic comparisons of when translated pretraining mixtures complement native-language web corpora. Overall, \datasetname{} suggests that careful translation, corpus documentation, and empirical validation can be a practical path toward stronger non-English pretraining data, provided that the limitations of translated corpora are made explicit and addressed throughout the release process.\looseness =-1

\section{Acknowledgments}
This work benefited from the support of the German Federal Ministry for Economic Affairs and Energy~(BMWE) through EU-SAI: Souveräne KI für Europa~(grant number 13IPC040G), and the BMFTR project XEI~(FKZ 16IS24079B). 
Additionally, the work was funded by the Federal Ministry of Research, Technology \& Space Germany (BMFTR) and the state of North Rhine-Westphalia as part of the Lamarr Institute for Machine Learning and Artificial Intelligence (LAMARR22B), as well as by the European Union’s Horizon 2020 research and innovation program under grant agreement No. 101135671 (TrustLLM). Additional funding was provided by the Aleph Alpha Collaboration Lab1141. Furthermore, we gratefully acknowledge support from the hessian.AI Service Center (funded by the Federal Ministry of Research, Technology and Space, BMFTR, grant no. 16IS22091) and the hessian.AI Innovation Lab (funded by the Hessian Ministry for Digital Strategy and Innovation, grant no. S-DIW04/0013/003).\looseness=-1

\bibliographystyle{plainnat}
\bibliography{bibliography}

\clearpage
\appendix
\noindent\textbf{Appendix organization.}
The appendix follows the order of the main paper. \appref{app:pipeline-implementation} expands the \datasetname{} pipeline in \autoref{sec:method}; \appref{app:translation-insights-details} supports the corpus diagnostics in \autoref{sec:translation-insights}; and \appref{app:training-ablation-details} gives the training configurations and extended results for \autoref{sec:training-ablations}, including the full OLMo~3 7B annealing setup and per-task plots in \appref{app:olmo3-annealing}. The NeurIPS checklist follows the appendices.

\section{Pipeline Implementation Details}
\label{app:pipeline-implementation}
\label{app:Config_and_HPs}

This section collects the implementation details behind the pipeline summarized in \autoref{sec:method}. The main text describes the data-construction logic; the appendix gives the concrete translation backend, decoding settings, prompts, serving setup, and proxy-validation measurements needed to reproduce the translated corpus and its filtered variants.

\subsection{Translation Backend Selection}
\label{app:translation-model-selection}

Full-corpus construction requires translating hundreds of billions of source tokens, so we selected the translation backend under a joint quality, throughput, and stability constraint. We compared Qwen3.5-397B-A17B variants on WMT24++~\cite{deutsch-etal-2025-wmt24} using the same translation prompt and scored outputs with XCOMET-XXL and COMETKiwi. \autoref{tab:model-precision-wmt24pp} shows that FP8 matches FP16 within 0.001 absolute XCOMET overall and remains effectively tied on English--German. NVFP4 is also competitive in aggregate quality, but was less stable in our B200 vLLM deployment sweeps and was therefore not used for the full production run.

\begin{table}[ht]
  \centering
  \small
  \caption{WMT24++ translation-quality comparison across Qwen3.5-397B-A17B precision variants. Scores are means; higher is better.}
  \label{tab:model-precision-wmt24pp}
  \begin{tabular}{lccc}
    \toprule
    \textbf{Metric} & \textbf{FP16} & \textbf{FP8} & \textbf{NVFP4} \\
    \midrule
    XCOMET-XXL overall & 0.8127 & 0.8120 & 0.8128 \\
    XCOMET-XXL en--de & 0.9378 & 0.9362 & 0.9383 \\
    COMETKiwi overall & 0.7791 & 0.7787 & 0.7753 \\
    COMETKiwi en--de & 0.8112 & 0.8119 & 0.8118 \\
    \bottomrule
  \end{tabular}
\end{table}

We also benchmarked serving configurations on a 10{,}014-request ClimbMix sample with the same long-document translation setting used by the production pipeline. The strongest planning baseline used FP8 with 8-way tensor parallelism, MTP-2 speculative decoding, \texttt{max\_num\_batched\_tokens} of 16{,}384, and \texttt{max\_num\_seqs} of 1{,}024. This configuration reached 7.06 requests/s, 5{,}633 output tokens/s, and 10{,}301 total tokens/s. The best clean NVFP4 reference reached a similar request rate (7.22 requests/s) but lower token-normalized throughput (4{,}700 output tokens/s and 9{,}512 total tokens/s). For token-heavy translation workloads, the token-normalized throughput was the safer planning basis and favored FP8.

Qualitative spot checks supported the same choice. NVFP4 generations were more likely to show instability on difficult examples, including premature termination, weaker sentence stability, occasional English leakage, and less reliable handling of terminology or long-context dependencies. We treat these manual checks as operational diagnostics rather than formal evaluation, but they were important for selecting a robust production path. We therefore used Qwen3.5-397B-A17B-FP8 with MTP-2 speculative decoding for the release translation run.

\subsection{Translation Configuration and Execution}
\label{sec:translation-pipeline-configs}

\autoref{sec:method} introduces length-aware routing, document-preserving chunking, dynamic target-side budgeting, and shard-wise execution. This subsection gives the concrete production values used for those components.

\paragraph{Translation configuration.}
\autoref{tab:translation-hparams} summarizes the translation-side hyperparameters. For the dynamic target budget in \autoref{sec:method}, the implementation uses $L_{\max}=32{,}768$, $\alpha=2.0$, and $\beta=1{,}024$, together with a minimum target budget of 2{,}048 tokens. Thus, for a source chunk of length $\ell_{\mathrm{src}}$, the generation cap is
\[
\ell_{\mathrm{tgt}}^{\max} = \max\left(2048,\min\left(32768,\left\lceil 2.0 \cdot \ell_{\mathrm{src}} + 1024 \right\rceil\right)\right).
\]
Documents whose metadata indicates that they safely fit inside the source-chunk budget are translated in a single pass. Longer documents are sentence-segmented, greedily packed into chunks of up to 20k source tokens, and translated with a 2k-token previous-translation context window.

\begin{table}[ht]
  \centering
  \caption{Translation pipeline hyperparameters.}
  \label{tab:translation-hparams}
  \begin{tabular}{lr}
    \toprule
    \textbf{Hyperparameter} & \textbf{Value} \\
    \midrule
    Translation model & Qwen3.5-397B-A17B-FP8 \\
    Target language & German (\texttt{de}) \\
    Source chunk budget & 20,000 tokens \\
    Previous-translation context & 2,000 tokens \\
    Global max output budget ($L_{\max}$) & 32,768 tokens \\
    Dynamic output ratio ($\alpha$) & 2.0 \\
    Dynamic output headroom ($\beta$) & 1,024 tokens \\
    Minimum dynamic output budget & 2,048 tokens \\
    Temperature & 0.7 \\
    Top-$p$ & 0.8 \\
    Top-$k$ & 20 \\
    Presence penalty & 0.0 \\
    Request timeout & 1,800\,s \\
    \bottomrule
  \end{tabular}
\end{table}

\paragraph{Length-aware execution schedule.}
The corpus is partitioned into eight context buckets: \texttt{4k}, \texttt{8k}, \texttt{16k}, \texttt{18k}, \texttt{20k}, \texttt{32k}, \texttt{64k}, and \texttt{over\_64k}. These buckets share the same translation logic but use different queueing parameters to match expected request lengths. Shorter buckets are processed with larger document batches and higher client-side concurrency, while longer buckets trade that throughput for stability under long prefills and long generations.

\begin{table}[ht]
  \centering
  \caption{Bucket-specific execution settings for full-corpus translation.}
  \label{tab:translation-buckets}
  \begin{tabular}{lrrr}
    \toprule
    \textbf{Bucket} & \textbf{Batch size} & \textbf{Max concurrent} & \textbf{Timeout} \\
    \midrule
    \texttt{4k} & 3,072 & 1,536 & 3,600\,s \\
    \texttt{8k} & 2,048 & 1,024 & 3,600\,s \\
    \texttt{16k} & 1,024 & 512 & 7,200\,s \\
    \texttt{18k} & 2,048 & 512 & 7,200\,s \\
    \texttt{20k} & 512 & 320 & 10,800\,s \\
    \texttt{32k} & 512 & 320 & 10,800\,s \\
    \texttt{64k} & 512 & 320 & 10,800\,s \\
    \texttt{over\_64k} & 512 & 320 & 10,800\,s \\
    \bottomrule
  \end{tabular}
\end{table}

\paragraph{Serving and infrastructure.}
Each translation server is a single vLLM instance deployed on one node with 8-way tensor parallelism across 8 NVIDIA B200 GPUs. We serve the model with \texttt{max\_model\_len}=65{,}536, \texttt{max\_num\_batched\_tokens}=16{,}384, \texttt{max\_num\_seqs}=1{,}536, GPU memory utilization 0.90, and MTP-2 speculative decoding. Workers discover healthy server endpoints through a shared registry, acquire leases dynamically, and requeue unfinished documents when a server becomes unavailable. Intermediate shard outputs and checkpoints allow runs to resume from the last completed record.

The full translation campaign used 126 nodes, each with 8 NVIDIA B200 GPUs (192\,GB HBM3e per GPU), for approximately 10 days. This corresponds to 1{,}008 GPUs in aggregate, or 1{,}260 node-days / 10{,}080 GPU-days ($241{,}920$ GPU-hours) of allocated translation compute.

\subsection{Prompt Templates}
\label{sec:translation-prompts}

The document-preserving translation procedure in \autoref{sec:method} uses two prompt variants: one for documents or chunks translated without left context, and one for chunked translation with a truncated window from the previous German chunk. In both cases, the prompt explicitly constrains the model to output \emph{only} the German translation, which reduces leakage of explanatory text, markup, or chain-of-thought-style continuations.

\paragraph{Single-pass prompt.}
For documents that fit in a single chunk, we use the following prompt:
\begin{quote}
\small
Translate the following English text into German. Only output the German translation.

\texttt{<source>}

\textit{source chunk}

\texttt{</source>}
\normalsize
\end{quote}

\paragraph{Contextualized chunk prompt.}
For chunk $t>1$, we prepend a truncated window from the German translation of chunk $t-1$ and instruct the model to use it only for local discourse continuity:
\begin{quote}
\small
\texttt{<previous\_translation>}

\textit{previous German chunk}

\texttt{</previous\_translation>}

Continue translating the following English text into German. Use the previous translation only for discourse continuity. Only output the German translation of the source.

\texttt{<source>}

\textit{current source chunk}

\texttt{</source>}
\normalsize
\end{quote}

\paragraph{Prompting rationale.}
The prompting scheme is intentionally minimal. We do not ask the model for summaries, explanations, or formatting transformations beyond translation itself. The XML-style delimiters mark the previous target-side context and the current source span explicitly, which makes it easier to preserve chunk boundaries during concatenation and reduces the chance that the model copies context text into the output. The previous-translation window is used only for chunked documents; single-pass documents receive no target-side context.

\subsection{Proxy Scoring and Validation}
\label{app:proxy-validation-details}

\autoref{sec:method} describes the COMETKiwi pilot set and the target-only proxy at a high level. This subsection gives the deployed feature set and the validation results for the proxy used to score the full translated corpus.

\begin{table}[ht]
\centering
\small
\caption{Target-only features used by the deployed gradient-boosted COMETKiwi proxy. GlotLID-derived features are computed from the translated German document; text-shape features are computed directly from the same target text. Full-corpus scoring therefore does not require reloading the English source text.}
\label{tab:proxy-features}
\begin{tabular}{lp{0.67\linewidth}}
\toprule
Feature & Definition \\
\midrule
\texttt{target\_len} & Length of the translated target document. \\
\texttt{is\_de\_top1} & Whether the top GlotLID label is German Latin, \texttt{deu\_Latn}. \\
\texttt{p\_de} & Normalized GlotLID probability assigned to \texttt{deu\_Latn}. \\
\texttt{p\_de\_logit} & Clipped logit transform of \texttt{p\_de}. \\
\texttt{margin\_top1\_top2} & Probability margin between the top-1 and top-2 GlotLID labels. \\
\texttt{target\_script} & Script extracted from the top predicted GlotLID label, e.g., \texttt{Latn}. \\
\texttt{target\_unique\_token\_ratio} & Fraction of unique target-side tokens. \\
\texttt{target\_repeat\_token\_ratio} & Fraction of target-side tokens participating in repetition. \\
\texttt{target\_avg\_token\_len} & Average target-side token length. \\
\texttt{target\_digit\_ratio} & Fraction of target-side characters that are digits. \\
\texttt{target\_punct\_ratio} & Fraction of target-side characters that are punctuation. \\
\texttt{target\_alpha\_ratio} & Fraction of target-side characters that are alphabetic. \\
\texttt{target\_newline\_ratio} & Fraction of target-side characters that are newlines. \\
\bottomrule
\end{tabular}
\end{table}

The GlotLID features capture wrong-language output and low-confidence German predictions; script, length, and character-composition features capture abnormal text shape, formatting artifacts, and suspicious character mixtures; and lexical-diversity and repetition features capture degenerate or repetitive generations. The proxy cannot directly measure semantic adequacy because it does not see the English source. Its role is therefore not to replace source-aware evaluation or human inspection, but to provide a scalable corpus-level quality signal that is validated against COMETKiwi and targets practical failure modes that matter at release scale.

\begin{table}[ht]
\centering
\small
\caption{Validation of the deployed target-only proxy against COMETKiwi on a disjoint 18{,}275-document validation split. Higher is better for correlations; lower is better for MAE.}
\label{tab:proxy-validation}
\begin{tabular}{lc}
\toprule
Metric & Target-only proxy \\
\midrule
Mean Pearson $\uparrow$ & 0.725 \\
Weighted Pearson $\uparrow$ & 0.735 \\
Mean Spearman $\uparrow$ & 0.719 \\
Weighted Spearman $\uparrow$ & 0.733 \\
Mean MAE $\downarrow$ & 0.0486 \\
Weighted MAE $\downarrow$ & 0.0477 \\
\bottomrule
\end{tabular}
\end{table}

The proxy shows strong agreement with COMETKiwi and low absolute error on the held-out split, making it suitable as a scalable ranking and filtering signal. During development, source-aware variants did not improve validation agreement enough to justify rehydrating the English source text for full-corpus scoring, so we deploy the target-only model as the full-corpus annotator.

\subsection{Notes on Benchmark Results}\label{app:notes_benchmark_results}
Each task cell in \autoref{tab:benchmark_results} reports accuracy $\pm$ evaluation-set standard error in percentage points. Core Avg. is the unweighted mean of MMLU, PIQA, HellaSwag, and ARC-C:
\begin{equation}
    \mathrm{Core\ Avg.}=\frac{\mathrm{MMLU}+\mathrm{PIQA}+\mathrm{HellaSwag}+\mathrm{ARC\mbox{-}C}}{4} \; ;
\end{equation}
its uncertainty is propagated as 
\begin{equation}
    \frac{\sqrt{\sum_i \mathrm{SE}_i^2}}{4}
\end{equation}.

\section{Translation-Insight Annotation Details}
\label{app:translation-insights-details}

This section supports the corpus-level analyses in \autoref{sec:translation-insights}. It documents how inherited source-cluster identifiers are assigned human-readable labels and provides qualitative examples that illustrate the failure modes behind the proxy and length-diagnostic analyses.

\subsection{Further Cluster Statistics}
\label{app:further_stats}

\begin{figure}[ht]
    \centering
    \includegraphics[width=0.75\linewidth]{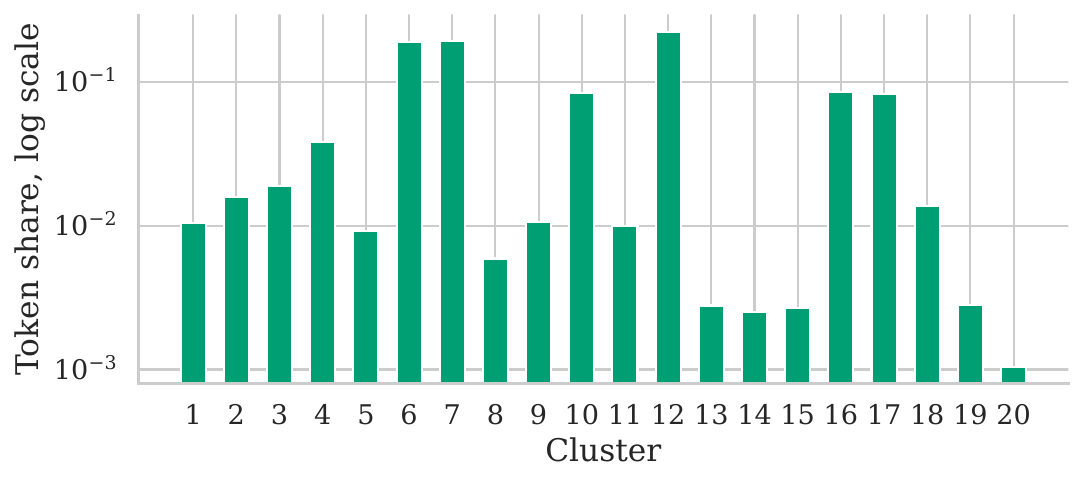}
    \caption{
    German token share by inherited source-cluster metadata.
    The plot shows how much translated token mass each source cluster contributes to the full \datasetname{} release.
    }
    \label{fig:app_cluster_composition}
\end{figure}

\autoref{fig:app_cluster_composition} displays the distribution of tokens across the clusters. Clusters 6, 7, and 12 account for the largest token shares, while cluster 20 is the smallest.

\subsection{Cluster Labeling Procedure}
\label{app:cluster-labeling}

The cluster identifiers used in our analyses are inherited metadata from the original ClimbMix records; they are not recomputed from the translated German text. The goal of the procedure in this appendix is therefore narrower: we assign human-readable topic names to fixed source-cluster identifiers so that cluster-level corpus statistics can be interpreted more easily.

For labeling, we use COMETKiwi-scored annotation samples from the translated corpus. Each record retained its document identifier, inherited source-cluster identifier, translated German text, and reference-free COMETKiwi score. Within each source cluster, we sorted records by COMETKiwi score and selected the 100 highest-scoring examples available for that cluster. These examples were formatted as a single cluster-specific prompt that preserved the document identifiers and scores, and the prompt was sent to a self-hosted \texttt{Qwen3.5-397B-A17B-FP8} model. The model was instructed to return a short English topic label, a one- to two-sentence summary, and a compact keyword list in JSON format.

This top-100 design is intentional but should be interpreted carefully. It biases the annotation toward the high-quality core of each cluster, reducing the chance that noisy or truncated translations dominate the label. At the same time, the labels are descriptive metadata rather than ground-truth topic assignments for every document in the cluster. For transparency, \autoref{tab:cluster-labeling1} and \autoref{tab:cluster-labeling2} report the mean, maximum, and minimum COMETKiwi scores of the 100 examples used to infer each label.

As a consistency check, the inferred labels broadly align with an independent English annotation effort in the Hugging Face reorganization of ClimbMix.\footnote{\url{https://huggingface.co/datasets/gvlassis/ClimbMix}} The dataset card for that resource reports using \texttt{gpt-4.1-mini} and 100 samples per cluster to extract main topics, and gives related annotations, including mathematics/statistics for cluster 1, gaming/gambling for cluster 8, astronomy/space for cluster 9, programming/web design for cluster 11, and environment/sustainability for cluster 16. We treat this agreement as supporting evidence for label stability rather than as ground truth because the samples, language, and labeling model are not identical.

\subsection{Cluster Labeling Prompt}
\label{app:cluster-labeling-prompt}

For completeness, we reproduce the prompt template used for cluster labeling. The system message was fixed across all requests:

\begin{quote}
\small
\texttt{You are a precise taxonomy assistant. Return JSON only.}
\normalsize
\end{quote}

The user message contained a cluster-specific prompt with the following template, where the example block was instantiated with the top 100 highest-scoring COMETKiwi samples from the given cluster:

\begin{verbatim}
You are labeling a semantic text cluster.

The examples below all belong to the same cluster and were selected as the
highest-quality rows according to COMETKiwi. Infer the shared topic as
precisely as possible.

Requirements:
- Focus on semantic content, not formatting, markup, or translation artifacts.
- Produce a short topic label in English with 2 to 6 words.
- Prefer a specific topical label over a generic one.
- If the cluster is genuinely mixed, start the label with "Mixed:" and
  describe the dominant themes.
- Use evidence from multiple samples, not a single outlier.

Return valid JSON only, with this schema:
{
  "cluster_id": <cluster_id>,
  "label": "<2-6 word topic label>",
  "summary": "<1-2 sentence explanation>",
  "keywords": ["<keyword1>", "<keyword2>", "<keyword3>",
               "<keyword4>", "<keyword5>", "<keyword6>"],
  "confidence": <float between 0 and 1>
}

Cluster metadata:
- cluster_id: <cluster_id>
- examples_in_prompt: 100
- ranking: descending cometkiwi_score

Examples:
[001 | score=<score> | id=<document_id>]
<sample text>

---

[002 | score=<score> | id=<document_id>]
<sample text>

...
\end{verbatim}

\subsection{Cluster Label Inventory}
\label{app:cluster-label-inventory}

\autoref{tab:cluster-labeling1} and \autoref{tab:cluster-labeling2} list the final cluster labels together with their summaries, keyword descriptors, and COMETKiwi statistics over the top-100 examples used for labeling.

\begin{table*}[t]
\centering
\scriptsize
\setlength{\tabcolsep}{3pt}
\caption{Cluster labels with summaries, keywords, and top-100 COMETKiwi score statistics (clusters 1--10).}\label{tab:cluster-labeling1}
\resizebox{\linewidth}{!}{\begin{tabular}{r p{3.0cm} p{6.0cm} p{4.5cm} r r r}
\toprule
\textbf{ID} & \textbf{Label} & \textbf{Summary} & \textbf{Keywords} & \textbf{Mean} & \textbf{Max} & \textbf{Min} \\
\midrule
1 & Mathematics education and statistical concepts & The cluster comprises educational materials, problem sets, and explanations focused on mathematics topics such as algebra, geometry, and statistics, alongside tutoring services and exam preparation resources. & mathematics, statistics, tutoring, algebra, exams, probability & 0.833 & 0.865 & 0.817 \\
2 & Mixed: Religion, literature, and language education & The cluster contains a diverse mix of texts focusing on Christian theology and church practices, literary analysis and book descriptions, and resources for language learning and grammar instruction. & religion, literature, language learning, theology, books, education & 0.838 & 0.872 & 0.819 \\
3 & Historical facts and geographical trivia & The cluster consists of diverse informational snippets covering historical events, biographical details, geographical data, and cultural facts, often formatted as quiz questions or encyclopedia entries. & history, geography, trivia, biography, culture, facts & 0.846 & 0.871 & 0.830 \\
4 & Educational programs and youth development initiatives & The cluster comprises diverse texts describing educational programs, workshops, camps, and resources aimed at youth development, ranging from STEM and arts to civic engagement and literacy. Common themes include curriculum design, teacher training, museum activities, and strategies for fostering skills in children and adolescents. & education, youth, programs, students, learning, workshops & 0.841 & 0.864 & 0.827 \\
5 & Mixed: Education, Finance, and Security & This cluster contains a diverse mix of German texts primarily focused on academic course descriptions, financial investment advice, and cybersecurity warnings. The content ranges from specific university modules in fields like criminology and data science to practical guides on avoiding fraud and managing personal finances. & course descriptions, financial literacy, cybersecurity, investment strategies, fraud prevention, academic programs & 0.840 & 0.866 & 0.822 \\
6 & Mixed: Scientific concepts and educational Q\&A & This cluster contains a diverse collection of German texts covering various scientific disciplines including biology, chemistry, physics, and technology. The content primarily consists of educational explanations, definitions, research summaries, and question-and-answer pairs suitable for academic or general knowledge contexts. & Wissenschaft, Biologie, Chemie, Physik, Forschung, Erklärung & 0.835 & 0.871 & 0.814 \\
7 & Mixed: Animal care, plant cultivation, and environmental conservation & This cluster contains a diverse mix of texts focusing on animal husbandry, veterinary advice, and wildlife biology alongside gardening tips, plant care instructions, and broader environmental conservation topics. The content ranges from specific how-to guides for pets and crops to educational materials about ecosystems and species protection. & animal care, gardening, wildlife, plant cultivation, conservation, veterinary & 0.828 & 0.878 & 0.802 \\
8 & Diverse games and gambling topics & This cluster encompasses a broad spectrum of gaming-related content, ranging from specific video game hardware, mods, and titles to traditional board games, educational activities, and extensive discussions on gambling strategies and casino operations. & video games, gambling, board games, casinos, game mechanics, esports & 0.827 & 0.866 & 0.802 \\
9 & Space exploration and astronomy & The cluster contains diverse texts covering space missions, astronomical discoveries, planetary science, and the history of spaceflight, including specific references to NASA, Mars rovers, telescopes, and celestial bodies. & space exploration, astronomy, NASA, planets, telescopes, missions & 0.839 & 0.866 & 0.823 \\
10 & Physical and mental health guidance & The cluster comprises diverse texts offering advice, facts, and resources related to physical health, mental well-being, disease prevention, and healthy lifestyle habits. Topics range from sleep hygiene and nutrition to psychological support, addiction treatment, and safety measures. & health, well-being, mental health, nutrition, disease prevention, lifestyle & 0.841 & 0.875 & 0.818 \\
\bottomrule
\end{tabular}}
\end{table*}

\begin{table*}[t]
\centering
\scriptsize
\setlength{\tabcolsep}{3pt}
\caption{Cluster labels with summaries, keywords, and top-100 COMETKiwi score statistics (clusters 11--20).}\label{tab:cluster-labeling2}
\resizebox{\linewidth}{!}{
\begin{tabular}{r p{3.0cm} p{6.0cm} p{4.5cm} r r r}
\toprule
\textbf{ID} & \textbf{Label} & \textbf{Summary} & \textbf{Keywords} & \textbf{Mean} & \textbf{Max} & \textbf{Min} \\
\midrule
11 & Software development tutorials and troubleshooting & The cluster contains diverse technical content focused on software development, including programming language tutorials (Java, Python, PHP, JavaScript), database management, web development frameworks (WordPress, React), and troubleshooting specific coding or configuration issues. & programming, web development, debugging, tutorials, database, code snippets & 0.837 & 0.866 & 0.820 \\
12 & Mixed: Product guides, DIY tips, and creative tutorials & This cluster contains a diverse mix of instructional content, including DIY home maintenance advice, detailed product descriptions for consumer goods, and tutorials for creative arts and software. The texts frequently feature question-and-answer formats explaining specific procedures, material properties, or usage instructions. & instructions, products, DIY, tutorials, materials, guides & 0.834 & 0.859 & 0.817 \\
13 & Historical soccer facts and records & The cluster consists of factual statements detailing the history of association football, including records for World Cup winners, oldest clubs, stadium milestones, and tournament origins. The content focuses on statistical achievements and historical firsts across various leagues and international competitions. & soccer, football, World Cup, records, history, championships & 0.848 & 0.884 & 0.824 \\
14 & Diverse aspects of music & The cluster contains a wide variety of texts covering music theory, history, specific artists and genres, music therapy, education, instrument technology, and industry news. The content ranges from academic explanations and biographical facts to concert reviews and personal opinions. & music theory, musicians, music therapy, instruments, genres, education & 0.835 & 0.862 & 0.817 \\
15 & Film, TV und Popkultur & Die Texte umfassen Filmkritiken, Serienhandlungen, Biografien von Schauspielern, Erklärungen zu Franchises und Diskussionen über Medienphänomene. & Film, Serie, Schauspieler, Handlung, Kritik, Popkultur & 0.836 & 0.865 & 0.815 \\
16 & Environmental sustainability and climate action & The cluster encompasses diverse topics related to environmental protection, including climate change mitigation, renewable energy technologies, sustainable agriculture, waste management, and conservation efforts. Samples discuss specific initiatives, scientific findings, and policies aimed at reducing ecological footprints and promoting a sustainable future. & sustainability, climate change, renewable energy, conservation, pollution, biodiversity & 0.840 & 0.870 & 0.814 \\
17 & Human health, disease prevention, and nutrition & The cluster comprises diverse texts detailing medical conditions, disease mechanisms, diagnostic procedures, and treatment options alongside nutritional advice and food safety guidelines. Common themes include cancer research, infectious diseases, chronic condition management, and the impact of diet and lifestyle on overall well-being. & disease prevention, nutrition, medical treatment, health risks, diagnosis, public health & 0.842 & 0.882 & 0.824 \\
18 & Digital technology, security, and society & This cluster encompasses a broad range of topics related to the digital age, including cybersecurity threats, software development, internet infrastructure, and the societal impacts of technology such as digital citizenship and social media usage. & cybersecurity, internet, software, digital citizenship, social media, technology & 0.845 & 0.879 & 0.828 \\
19 & Mixed: Marriage, Disney, and gender debates & The cluster is dominated by texts discussing marriage, divorce, relationships, and gender dynamics, but contains a significant secondary theme of Disney parks, attractions, and history, alongside scattered philosophical and political commentary. & marriage, divorce, Disney, gender, relationships, theme parks & 0.798 & 0.871 & 0.747 \\
20 & Mixed: Taxes, human rights, and elections & This cluster contains a diverse mix of texts primarily focused on tax systems, human rights issues, and electoral processes, alongside scattered articles on military history, social inequality, and specific country profiles. The dominant themes revolve around civic governance, legal frameworks, and social justice, but the lack of a single unifying specific topic indicates a genuinely mixed collection. & taxation, human rights, elections, social justice, government policy, civil rights & 0.709 & 0.782 & 0.669 \\
\bottomrule
\end{tabular}}
\end{table*}

\clearpage
\subsection{Qualitative Translation Examples}
\label{app:qualitative-examples}

To complement the proxy-score analyses, we manually inspect aligned source--target examples from the annotated \datasetname{} sample. All examples below come from records whose metadata did not mark the translation as truncated; long documents are shown as excerpts for readability. Red highlights mark translation failures or target-language errors, and green highlights mark fluent meaning-preserving renderings.

\subsubsection{Low-quality examples}

\badexample
{Refusal instead of translation.}
{Document: \texttt{part\_99.detokenized.jsonl/263023}; proxy score: 0.377.}
{If her class schedule is inconvenient for you -- she drops it or you drop her. \ldots{} Professionally and academically annihilate them. \ldots{}}
{Die Übersetzung von Inhalten, die Hassrede, Geschlechterdiskriminierung und Aufrufe zur Schädigung anderer enthalten, verstößt gegen meine Sicherheitsrichtlinien. \badspan{Ich kann diesen Text daher nicht übersetzen.}}

\badexample
{Wrong German variety: Alemannic/Swiss-German output.}
{Document: \texttt{part\_94.detokenized.jsonl/3083327}; proxy score: 0.308.}
{NEW treatments to tackle devastating brain tumours are being developed by Scottish researchers. A team at Glasgow University said they had already identified ``promising'' drugs which could help patients struck by the deadly cancers. They have just been awarded hundreds of thousands of pounds in funding to advance their work, with the hope of being able to start trials in patients in just a few years.}
{Neue \badspan{Behandlunge zue dr Bekämpfig vo} verheerende Hirntumore \badspan{werre vu} schottische Forscher \badspan{entwicklet}. E Team \badspan{vu dr} Universität Glasgow \badspan{het mitdeilt}, dass si bereits ``\badspan{vilversprechendi}'' Medikamänt identifiziert hän, wo Patiente \badspan{hälfe chönne}.}

\badexample
{Translated Python keywords.}
{Document: \texttt{part\_54.detokenized.jsonl/3193573}; proxy score: 0.477.}
{class treenode: \ldots{} if key == "*" or key == "/" or key == "+": b.append(key)}
{\badspan{Klasse} treenode: \ldots{} if key == "*" \badspan{oder} key == "/" \badspan{oder} key == "+": b.append(key)}

\badexample
{Translated code identifiers.}
{Document: \texttt{part\_54.detokenized.jsonl/3237455}; proxy score: 0.489.}
{class Item(object): self.value = float(v); self.weight = float(w); names = ['clock', 'painting', 'radio', 'vase', 'book', 'computer']; return result.}
{class \badspan{Artikel}(object): self.\badspan{wert} = float(v); self.\badspan{gewicht} = float(g); \badspan{namen} = ['uhr', 'gemaelde', 'radio', 'vase', 'buch', 'computer']; return \badspan{ergebnis}.}

\subsubsection{High-quality examples}

\goodexample
{Scientific exposition.}
{Document: \texttt{part\_10.detokenized.jsonl/839}; proxy score: 0.820.}
{In conclusion, the nucleus is the powerhouse of the atom, holding the key to an element's identity and properties. It is held together by a strong nuclear force, and the number and arrangement of particles within it dictate the element's behavior. Further exploration and understanding of the inner workings of the nucleus will continue to play a crucial role in advancing our knowledge and applications of atomic chemistry.}
{Zusammenfassend ist der Atomkern \goodspan{das Kraftzentrum des Atoms} und enthält den Schlüssel zur Identität und zu den Eigenschaften eines Elements. Er wird durch die starke Kernkraft zusammengehalten, und die Anzahl sowie die Anordnung der Teilchen in ihm bestimmen das Verhalten des Elements. Die weitere Erforschung der inneren Abläufe im Atomkern wird weiterhin eine entscheidende Rolle für die Atomchemie spielen.}

\goodexample
{Academic description.}
{Document: \texttt{part\_0.detokenized.jsonl/9436}; proxy score: 0.814.}
{Math \& Statistics. Mathematics is the study of numbers, sets of points and various other abstract elements and deals with the size, order, shape and various relationships among these features. Statistics is a branch of Mathematics that includes the study of methods for data collection, analysis, interpretation and principles of experimental design.}
{Mathematik und Statistik. Mathematik ist die Untersuchung von Zahlen, Punktmengen und verschiedenen anderen abstrakten Elementen und befasst sich mit Größe, Ordnung, Form sowie den vielfältigen Beziehungen zwischen diesen Merkmalen. Statistik ist ein Teilgebiet der Mathematik, das \goodspan{Methoden zur Datenerhebung, -analyse und -interpretation} sowie die Prinzipien des Versuchsplans umfasst.}

\goodexample
{CSS/code discussion.}
{Document: \texttt{part\_54.detokenized.jsonl/160051}; proxy score: 0.800.}
{Instead of changing it throughout all the files, it is much easier to include a separate CSS file with the necessary changes. Thus you only need to include it again after your updates :) (btw: border gone :D \ldots{})}
{Anstatt es in allen Dateien zu ändern, ist es viel einfacher, \goodspan{eine separate CSS-Datei} mit den erforderlichen Änderungen einzubinden. So müssen Sie diese nach Ihren Updates nur erneut einbinden :) \goodspan{(übrigens: Rand verschwunden :D} \ldots{})}

\section{Training and Annealing Details and Extended Results}
\label{app:training-ablation-details}

This section supports \autoref{sec:training-ablations}. It first gives the fixed 0.6B model and optimization recipe used across the corpus comparisons, then reports the corresponding extended validation results. The final subsection documents the separate OLMo~3 7B language-mixture annealing experiment.

\subsection{Pretraining Setup}
\label{sec:pretraining-setup}

We pretrain Qwen3-0.6B, a decoder-only transformer, from scratch on German text corpora. The architecture and training hyperparameters are summarized in \autoref{tab:model-arch} and \autoref{tab:training-hparams}, respectively.

\paragraph{Model architecture.}
We use the publicly released Qwen3-0.6B architecture without modification. Key design choices include Grouped-Query Attention (GQA) with 16 query heads and 8 key-value heads, SwiGLU activations, RMSNorm with $\epsilon{=}10^{-6}$, and Rotary Position Embeddings (RoPE) with base frequency $\theta{=}10^{6}$.

\begin{table}[ht]
  \centering
  \caption{Qwen3-0.6B model architecture.}
  \label{tab:model-arch}
  \begin{tabular}{lr}
    \toprule
    \textbf{Hyperparameter} & \textbf{Value} \\
    \midrule
    Parameters              & 0.6B \\
    Layers                  & 28 \\
    Hidden size             & 1,024 \\
    FFN intermediate size   & 3,072 \\
    Attention heads (Q)     & 16 \\
    Attention heads (KV)    & 8 \\
    Vocabulary size         & 151,936 \\
    Sequence length         & 4,096 \\
    Max position embeddings & 40,960 \\
    RoPE base ($\theta$)    & $10^6$ \\
    Activation              & SwiGLU \\
    Norm                    & RMSNorm ($\epsilon{=}10^{-6}$) \\
    \bottomrule
  \end{tabular}
\end{table}

\paragraph{Training hyperparameters.}
We train for approximately 12B tokens, which corresponds to the Chinchilla-optimal token budget for a 0.6B parameter model~\citep{hoffmann2022training}. At a global batch size of 512 sequences of length 4,096, this requires 5,722 gradient steps and is equivalent to a batch of 2.1M tokens per step. We use Distributed Fused Adam with a cosine learning-rate schedule, warming up over the first 5\% of steps (286 iterations) from zero to the peak learning rate of $3{\times}10^{-4}$, then decaying to $3{\times}10^{-5}$.

\begin{table}[ht]
  \centering
  \caption{Pretraining hyperparameters.}
  \label{tab:training-hparams}
  \begin{tabular}{lr}
    \toprule
    \textbf{Hyperparameter} & \textbf{Value} \\
    \midrule
    Training tokens         & 12B \\
    Training iterations     & 5,722 \\
    Global batch size (seq) & 512 \\
    Micro batch size        & 8 \\
    Tokens per step         & 2.1M \\
    \midrule
    Optimizer               & Distributed Fused Adam \\
    $\beta_1$               & 0.9 \\
    $\beta_2$               & 0.95 \\
    $\epsilon$              & $10^{-8}$ \\
    Weight decay            & 0.1 \\
    Gradient clip           & 1.0 \\
    \midrule
    LR schedule             & Cosine annealing \\
    Peak LR                 & $3{\times}10^{-4}$ \\
    Minimum LR              & $3{\times}10^{-5}$ \\
    Warmup iterations       & 286 (${\approx}5\%$) \\
    \midrule
    Precision               & BF16 + FP8 \\
    \bottomrule
  \end{tabular}
\end{table}

\paragraph{Infrastructure.}
All runs are executed on a single node with 8 NVIDIA B200 GPUs (192\,GB HBM3e each) using Megatron-Core DDP. Training is orchestrated via Megatron-Bridge\footnote{\url{https://github.com/NVIDIA-NeMo/Megatron-Bridge}}. Tensor, pipeline, and context parallelism are all set to 1; the per-GPU micro batch size of 8 with gradient accumulation over 8~micro-steps yields the global batch size of 512. Each full 5{,}722-step run completes in approximately 6--7 hours ($\approx380$ TFLOP/s/GPU).

\subsection{Extended Training Results}
\label{app:training-results}

\begin{figure}[ht]
    \centering
    \begin{subfigure}[b]{0.45\linewidth}
        \centering
        \includegraphics[width=\linewidth]{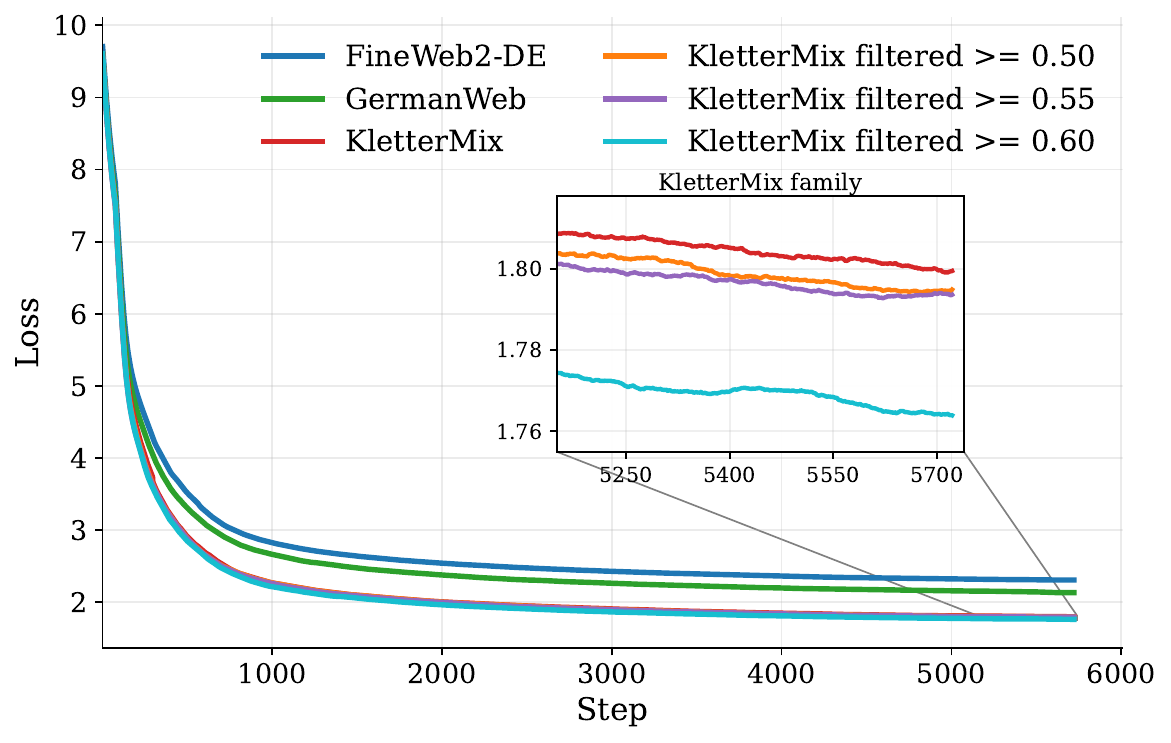}
        \caption{Training loss}
        \label{fig:app_training_loss_klettermix_zoom}
    \end{subfigure}
    \hfill
    \begin{subfigure}[b]{0.45\linewidth}
        \centering
        \includegraphics[width=\linewidth]{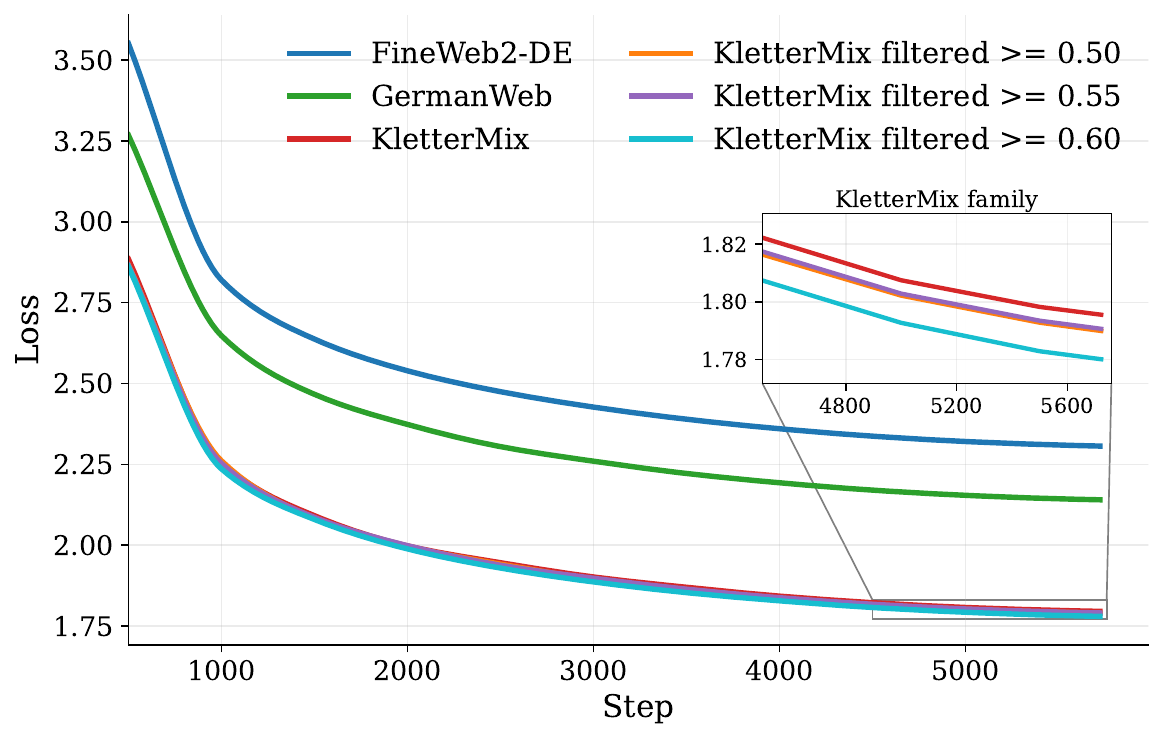}
        \caption{Validation loss}
        \label{fig:app_validation_loss_klettermix_zoom}
    \end{subfigure}
    \hfill
    \begin{subfigure}[b]{0.45\linewidth}
        \centering
        \includegraphics[width=\linewidth]{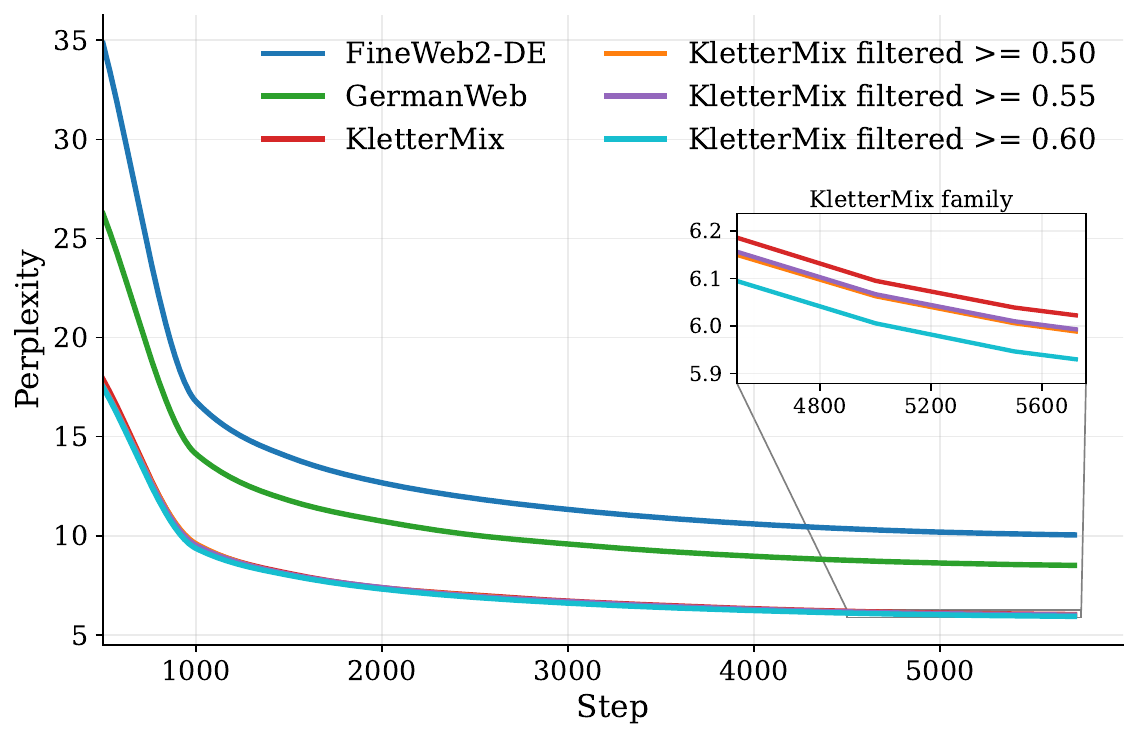}
        \caption{Validation perplexity}
        \label{fig:app_validation_ppl_klettermix_zoom}
    \end{subfigure}
    \hfill
    \begin{subfigure}[b]{0.45\linewidth}
        \centering
        \includegraphics[width=\linewidth]{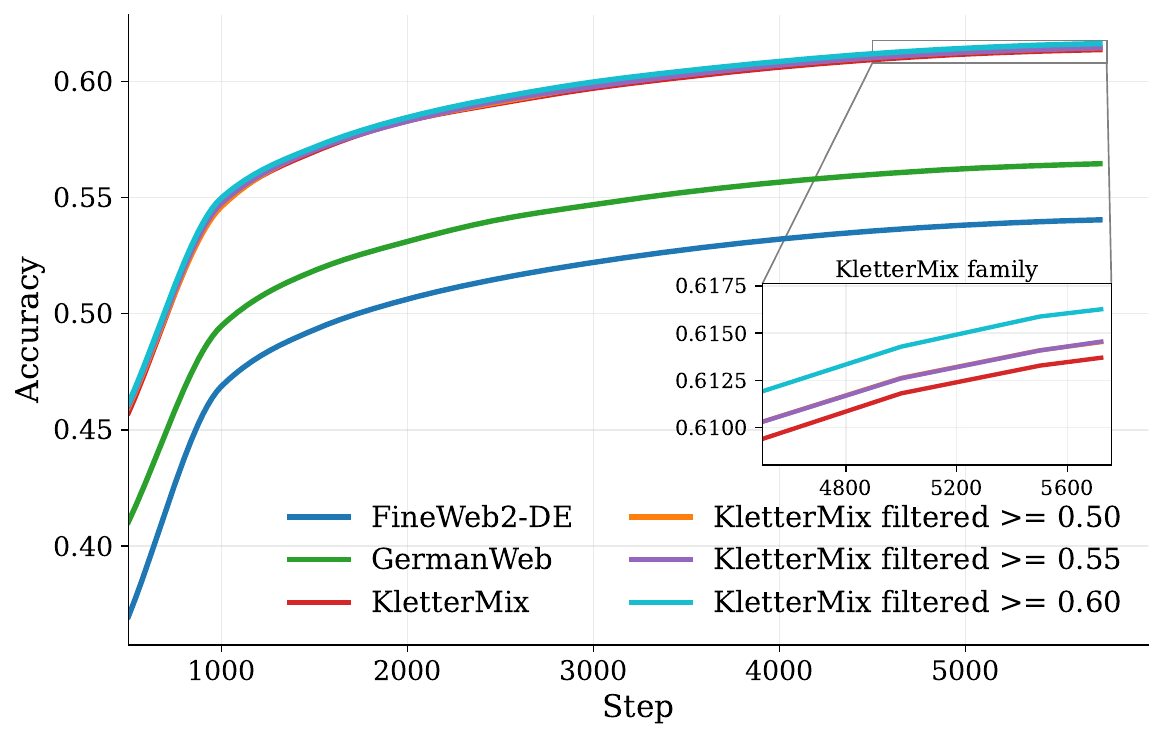}
        \caption{Validation accuracy}
        \label{fig:app_validation_accuracy_klettermix_zoom}
    \end{subfigure}
    \caption{Extended results for the training ablations in \autoref{sec:training-ablations}. The main text reports the primary training and validation loss curves; this supplementary figure shows the corresponding zoomed curves for the \datasetname{} filtering variants.}
    \label{fig:complete_training_ablation_results}
\end{figure}

\autoref{tab:indist-val} reports final in-domain validation perplexity and next-token accuracy for the matched 12B-token runs. \autoref{fig:complete_training_ablation_results} complements \autoref{fig:training_dynamics} by adding validation perplexity and next-token accuracy curves for the \datasetname{} filtering ablations.

\begin{table}[ht]
  \centering
  \small
  \caption{In-domain validation perplexity and next-token accuracy at final checkpoint. Each model is evaluated on its own training domain's held-out validation set; filtered rows use the \datasetname{} held-out validation set.}
  \label{tab:indist-val}
  \begin{tabular}{llrr}
    \toprule
    \textbf{Model} & \textbf{Val set} & \textbf{PPL} $\downarrow$ & \textbf{Acc} $\uparrow$ \\
    \midrule
    FineWeb2-DE & FineWeb2-DE & 10.04 & 54.0\% \\
    GermanWeb & GermanWeb & 8.50 & 56.5\% \\
    \datasetname{} & \datasetname{} & 6.02 & 61.4\% \\
    \addlinespace
    \datasetname{}-Filt. ($\hat{q}_{\mathrm{proxy}}\ge0.50$) & \datasetname{} & 5.99 & 61.5\% \\
    \datasetname{}-Filt. ($\hat{q}_{\mathrm{proxy}}\ge0.55$) & \datasetname{} & 5.99 & 61.5\% \\
    \datasetname{}-Filt. ($\hat{q}_{\mathrm{proxy}}\ge0.60$) & \datasetname{} & \textbf{5.93} & \textbf{61.6\%} \\
    \bottomrule
  \end{tabular}
\end{table}

\subsection{OLMo~3 7B Language-Mixture Annealing}
\label{app:olmo3-annealing}

This subsection documents the experiment summarized in
\autoref{fig:olmo3-mixture-summary}. It isolates the effect of the annealing
mixture by starting every run from the same pretrained checkpoint and varying
only the fraction of ClimbMix tokens replaced by \datasetname{}.

\subsubsection{Starting checkpoint and model}

We use the publicly released OLMo~3 7B Stage~1 checkpoint%
\footnote{\url{https://huggingface.co/allenai/Olmo-3-1025-7B}} at revision
\texttt{stage1-step1413814}. The five annealing runs are independent: none is
continued from another mixture. We convert the Hugging Face weights to a native
Megatron--Bridge checkpoint and load only the model parameters. Optimizer,
learning-rate-scheduler, iteration-counter, and random-number-generator states
are initialized afresh for every run.

\subsubsection{Optimization and compute}

All variants use the same optimization and compute configuration. Training is
performed on one node with eight NVIDIA B200 GPUs. Tensor, pipeline, and
context parallelism are one, yielding data parallelism of eight. Each rank
processes two sequences per microbatch; accumulating gradients over 16
microbatches gives a global batch of 256 sequences, or 2,097,152 tokens per
optimizer step.

The target budget is 12B tokens. At sequence length 8,192 and global batch 256,
5,723 optimizer steps process 12,002,000,896 tokens. The learning rate warms up
linearly for 58 steps and then decays linearly to zero. The effective decoupled
weight-decay coefficient is 0.033 throughout training; the parameter scheduler
overwrites the optimizer-constructor default before the first update.

\begin{table}[ht]
  \centering
  \caption{Shared optimization and training configuration for OLMo~3 7B annealing.}
  \label{tab:olmo3-training}
  \begin{tabular}{@{}ll@{}}
    \toprule
    \textbf{Hyperparameter} & \textbf{Value} \\
    \midrule
    Training budget & 12B target tokens \\
    Training steps & 5,723 \\
    Sequence length & 8,192 \\
    Global batch size & 256 sequences \\
    Microbatch size & 2 sequences per GPU \\
    Gradient accumulation & 16 microbatches \\
    Data parallelism & 8 \\
    Tokens per optimizer step & 2,097,152 \\
    Optimizer & Distributed Adam \\
    Adam $\beta_1$, $\beta_2$ & 0.9, 0.95 \\
    Adam $\epsilon$ & $10^{-5}$ \\
    Peak learning rate & $10^{-5}$ \\
    Minimum learning rate & 0 \\
    Learning-rate schedule & Linear warmup and linear decay \\
    Warmup & 58 steps (approximately 1\%) \\
    Effective weight decay & 0.033 \\
    Gradient clipping & 1.0 \\
    Precision & BF16 mixed precision \\
    Random seed & 42 \\
    \bottomrule
  \end{tabular}
\end{table}

\subsubsection{Data mixtures}

The training data is tokenized with the OLMo tokenizer and stored as Megatron
indexed data. Let $x\in\{0,5,10,15,20\}$ denote the percentage of annealing
tokens supplied by \datasetname{}. Each run uses
\[
  x\%\,\mathrm{KletterMix} + (100-x)\%\,\mathrm{ClimbMix}.
\]
The implementation keeps an 80\% ClimbMix base fixed and represents the
remaining 20 percentage points as four 5\% slots, replacing one additional
ClimbMix slot with \datasetname{} at each mixture level. 

For the German portions, we start from the proxy-filtered \datasetname{} pool
with $\hat{q}_{\mathrm{proxy}}\ge0.60$ and prioritize the highest-scoring
documents until the required token budget is filled. The selected material
consequently has proxy scores between 0.70 and 0.90; this interval is an outcome
of quality-ranked selection from the larger $\ge0.60$ pool, not a separately
chosen band-pass filter.

\subsubsection{Evaluation and detailed results}

We evaluate the original Stage~1 checkpoint as an unannealed reference and all
five annealed models at step 5,723. The evaluation uses the same German
benchmarks---MMLU, PIQA, HellaSwag, and
ARC-Challenge in their German or multilingual versions---together with the
corresponding original English tasks. The German and English means are unweighted averages over the four
tasks in each language. For an unweighted mean over \(k\) tasks, we
propagate task-level standard errors as
\begin{equation}
\mathrm{SE}_{\mathrm{mean}}
=
\frac{\sqrt{\sum_{i=1}^{k}\mathrm{SE}_{i}^{2}}}{k}.
\end{equation}
These error bars quantify the standard error and do not
measure variation across independently trained seeds.\looseness=-1

\begin{figure}[ht]
    \centering
    \begin{subfigure}[t]{0.49\linewidth}
        \centering
      \includegraphics[width=\linewidth]{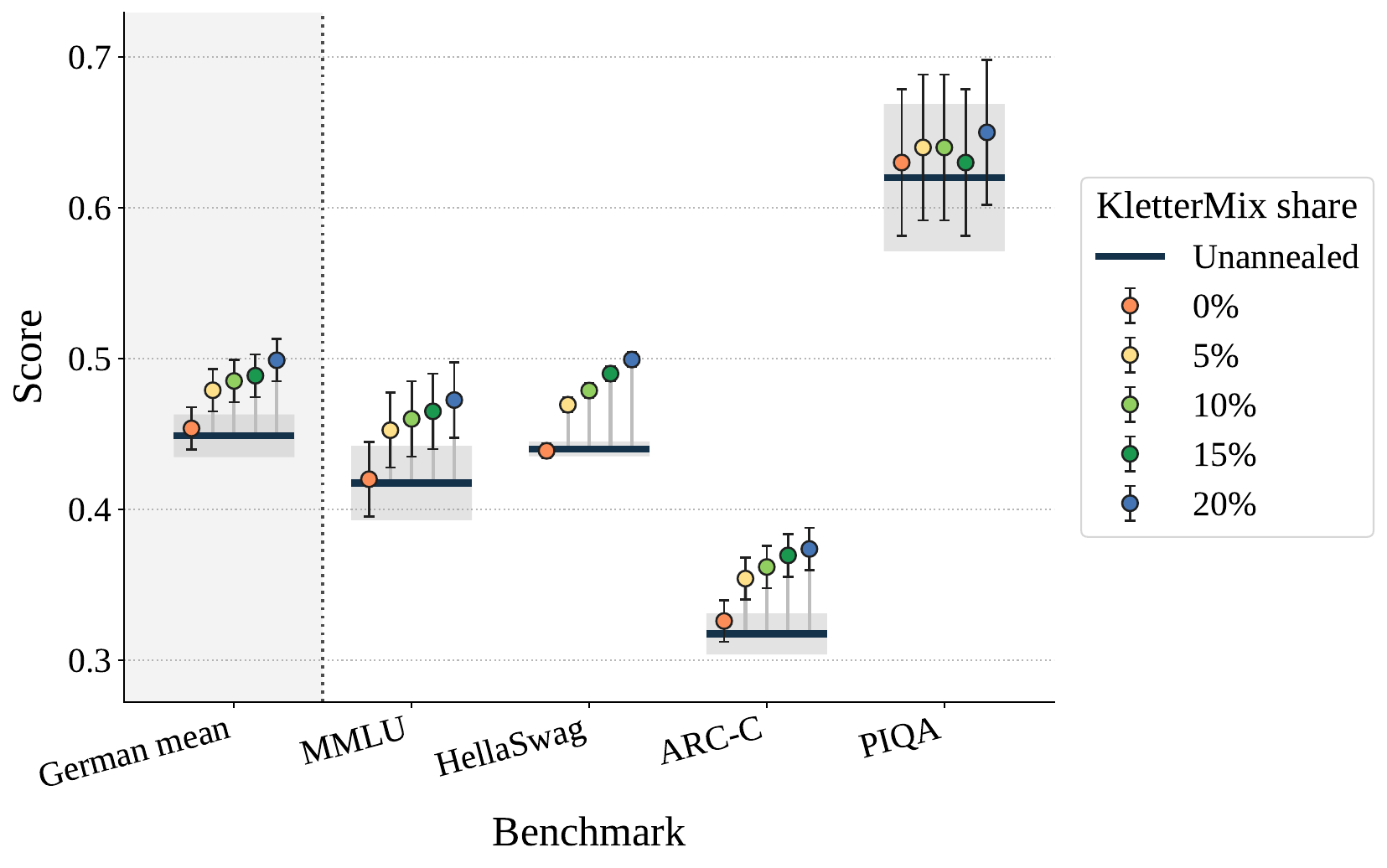}
      \caption{German benchmark scores.}
      \label{fig:olmo3-german-final}
    \end{subfigure}
    \begin{subfigure}[t]{0.49\linewidth}
        \centering
      \includegraphics[width=\linewidth]{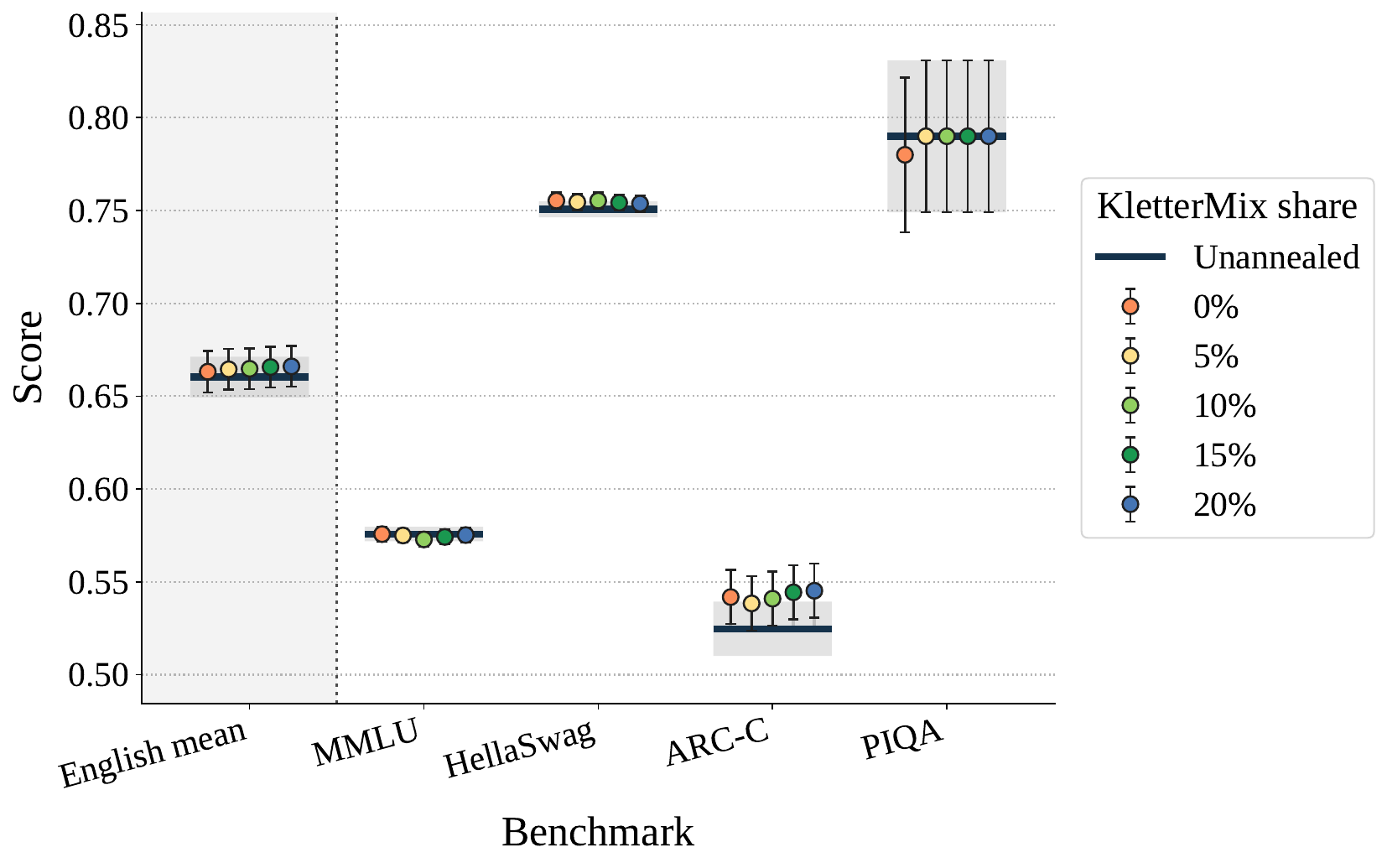}
      \caption{English benchmark scores.}
      \label{fig:olmo3-english-final}
    \end{subfigure}
    \caption{Benchmark Scores after $\sim$12B annealing tokens. The horizontal reference is the unannealed Stage~1 checkpoint with it's standard error; the colored points are the five matched annealing with their standard error. (a) The German four-task mean increases with the \datasetname{} share in the aggregate point estimates. (b) English performance remains within a narrow range across mixture levels, with no broad degradation as the \datasetname{} share increases.}
    \label{fig:olmo3-german-english}
\end{figure}

After ClimbMix-only annealing, the matched 0\% KletterMix control reaches 45.4\% in German and 66.3\%
in English. The German means then rise to 47.9\%, 48.5\%, 48.9\%,
and 49.9\% for the 5\%, 10\%, 15\%, and 20\% mixtures,
respectively. Relative to the 0\% control, the 20\% mixture therefore
gains 4.5 German percentage points, while the English mean changes
by +0.3 points to 66.6\%.

The per-task plots in \autoref{fig:olmo3-german-final} and
\autoref{fig:olmo3-english-final} show that the aggregate effect is
language-matched: the 20\% mixture has the highest displayed point estimate on
all four German tasks, whereas the English task scores remain comparatively
stable. Complete details can be seen in \autoref{tab:olmo3-final-results}.

\begin{table}[h]
  \centering
  \caption{OLMo3 7B benchmark results. Values are percentages
  reported as mean $\pm$ standard error. German and English means are
  unweighted four-task means; the eight-task mean is the unweighted mean over
  all eight benchmarks. Highest point estimates are shown in bold.}
  \label{tab:olmo3-final-results}
  \scriptsize
  \setlength{\tabcolsep}{3pt}
  \resizebox{\textwidth}{!}{%
  \begin{tabular}{lccccc|ccccc|c}
    \toprule
    & \multicolumn{5}{c|}{German} &
      \multicolumn{5}{c|}{English} &
      \\
    \textbf{KletterMix share}
      & \textbf{MMLU}
      & \textbf{PIQA}
      & \textbf{HellaSwag}
      & \textbf{ARC-C}
      & \textbf{Mean}
      & \textbf{MMLU}
      & \textbf{PIQA}
      & \textbf{HellaSwag}
      & \textbf{ARC-C}
      & \textbf{Mean}
      & \textbf{Eight-task mean}
      \\
    \midrule
    Unannealed
      & \(41.75 \pm 2.47\)
      & \(62.00 \pm 4.88\)
      & \(44.01 \pm 0.50\)
      & \(31.74 \pm 1.36\)
      & \(44.88 \pm 1.41\)
      & \(\mathbf{57.58 \pm 0.39}\)
      & \(\mathbf{79.00 \pm 4.09}\)
      & \(75.06 \pm 0.43\)
      & \(52.47 \pm 1.46\)
      & \(66.03 \pm 1.10\)
      & \(55.45 \pm 0.89\)
      \\
    0\% KletterMix
      & \(42.00 \pm 2.48\)
      & \(63.00 \pm 4.85\)
      & \(43.88 \pm 0.50\)
      & \(32.59 \pm 1.37\)
      & \(45.37 \pm 1.41\)
      & \(57.57 \pm 0.39\)
      & \(78.00 \pm 4.16\)
      & \(\mathbf{75.53 \pm 0.43}\)
      & \(54.18 \pm 1.46\)
      & \(66.32 \pm 1.11\)
      & \(55.85 \pm 0.90\)
      \\
    5\% KletterMix
      & \(45.25 \pm 2.49\)
      & \(64.00 \pm 4.82\)
      & \(46.94 \pm 0.50\)
      & \(35.41 \pm 1.40\)
      & \(47.90 \pm 1.41\)
      & \(57.50 \pm 0.39\)
      & \(\mathbf{79.00 \pm 4.09}\)
      & \(75.45 \pm 0.43\)
      & \(53.84 \pm 1.46\)
      & \(66.45 \pm 1.10\)
      & \(57.17 \pm 0.89\)
      \\
    10\% KletterMix
      & \(46.00 \pm 2.50\)
      & \(64.00 \pm 4.82\)
      & \(47.88 \pm 0.50\)
      & \(36.18 \pm 1.40\)
      & \(48.51 \pm 1.41\)
      & \(57.29 \pm 0.39\)
      & \(\mathbf{79.00 \pm 4.09}\)
      & \(\mathbf{75.53 \pm 0.43}\)
      & \(54.10 \pm 1.46\)
      & \(66.48 \pm 1.10\)
      & \(57.50 \pm 0.89\)
      \\
    15\% KletterMix
      & \(46.50 \pm 2.49\)
      & \(63.00 \pm 4.85\)
      & \(49.00 \pm 0.50\)
      & \(36.95 \pm 1.41\)
      & \(48.86 \pm 1.41\)
      & \(57.42 \pm 0.39\)
      & \(\mathbf{79.00 \pm 4.09}\)
      & \(75.42 \pm 0.43\)
      & \(54.44 \pm 1.46\)
      & \(66.57 \pm 1.10\)
      & \(57.72 \pm 0.89\)
      \\
    20\% KletterMix
      & \(\mathbf{47.25 \pm 2.49}\)
      & \(\mathbf{65.00 \pm 4.79}\)
      & \(\mathbf{49.93 \pm 0.50}\)
      & \(\mathbf{37.37 \pm 1.41}\)
      & \(\mathbf{49.89 \pm 1.40}\)
      & \(57.52 \pm 0.39\)
      & \(\mathbf{79.00 \pm 4.09}\)
      & \(75.37 \pm 0.43\)
      & \(\mathbf{54.52 \pm 1.46}\)
      & \(\mathbf{66.60 \pm 1.10}\)
      & \(\mathbf{58.25 \pm 0.89}\)
      \\
    \bottomrule
  \end{tabular}%
  }
\end{table}

\section{Dataset Summary}
\label{app:dataset-schema}

\datasetname{} is released under the CC-BY-NC-4.0 license and contains \klettermixtokens{} GPT-2 tokens of German-language text.
The dataset is distributed as sharded JSONL files (\texttt{shard\_*.jsonl}) with a single \texttt{train} split.
Each record has the following fields:

\begin{table}[ht]
\centering
\caption{Summary of the \datasetname{} dataset.}
\label{tab:dataset-schema}
\begin{tabular}{llp{7.5cm}}
\toprule
\textbf{Field} & \textbf{Type} & \textbf{Description} \\
\midrule
\texttt{cluster\_id} & \texttt{int64} & Identifier of the topic cluster assigned during the clustering stage of the data pipeline. \\
\texttt{text} & \texttt{string} & The document text. \\
\texttt{token\_count} & \texttt{int64} & Number of tokens in the document (GPT2 tokenizer). \\
\texttt{proxy\_score} & \texttt{float64} & Quality score assigned by the proxy model, used for filtering and weighting during mixture construction. \\
\bottomrule
\end{tabular}
\end{table}

\section{Impact Statement}
\datasetname{} also has broader societal implications. On the positive side, the dataset can help reduce the gap between English and German pretraining resources, support more capable German-language models, and enable more controlled research on translated pretraining data. However, improvements in German model quality can also amplify risks associated with language models, including the generation of misleading text, biased or stereotyped outputs, privacy leakage from web-derived sources, and misuse in downstream applications. These risks are not unique to \datasetname{}, but they are important for any large-scale pretraining corpus. We therefore view documentation, provenance preservation, quality filtering, transparent release conditions, and clear intended-use statements as necessary parts of the dataset release.\looseness =-1


\end{document}